\documentclass[10pt,twocolumn,letterpaper]{article}

\usepackage[pagenumbers]{wacv} 

\usepackage{graphicx}
\usepackage{amsmath}
\usepackage{amssymb}
\usepackage{booktabs}
\usepackage{comment}
\usepackage{bm}
\usepackage{pifont}
\usepackage{caption}
\usepackage{cuted}

\usepackage[pagebackref,breaklinks,colorlinks]{hyperref}

\usepackage[accsupp]{axessibility}

\usepackage[capitalize]{cleveref}
\crefname{section}{Sec.}{Secs.}
\Crefname{section}{Section}{Sections}
\Crefname{table}{Table}{Tables}
\crefname{table}{Tab.}{Tabs.}

\AtBeginDocument{
  \setlength\abovedisplayskip{0pt}
  \setlength\belowdisplayskip{0pt}}
\usepackage[margin=0pt, font=small]{caption}
\usepackage[activate={true,nocompatibility},final,tracking=false,kerning=true,spacing=true,factor=1100,stretch=10,shrink=20]{microtype}
\usepackage[moderate]{savetrees}

\def\wacvPaperID{1884} 
\def\confName{WACV}
\def\confYear{2025}

\begin{document}

\title{Lorentz Framework for Semantic Segmentation}


\let\oldtwocolumn\twocolumn
\renewcommand\twocolumn[1][]{%
    \oldtwocolumn[{#1}{
    \vspace{-3.5em}
    \begin{center}
        \includegraphics[width=0.95\textwidth]{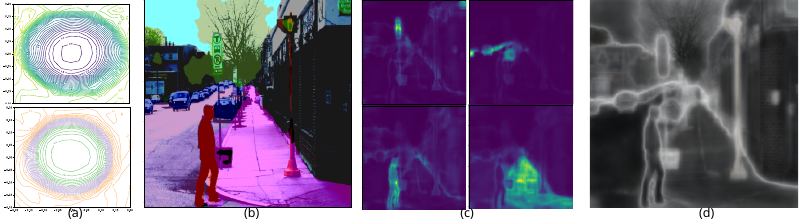}
        \captionof{figure}{(a) Lorentz model achieves flatter minima (bottom) compared to Euclidean model (top). (b) Semantic mask for a sample of ADE20K (c) Confidence map generated for traffic sign, cars, person, and sidewalk (c) Boundary Estimation by minimum exterior angle. } 
        \label{fig:intro}
        \end{center}
    }]
}


\author{
Zahid Hasan\thanks{Corresponding author:  \href{mailto:zhasan3@umbc.edu}{zhasan3@umbc.edu}} \quad Masud Ahmed \quad Nirmalya Roy \\
Department of Information Systems, University of Maryland, Baltimore County\\
{\tt\small \{zhasan3, mahmed10, nroy\}@umbc.edu}
}

\maketitle
\begin{abstract}
Semantic segmentation in hyperbolic space enables compact modeling of hierarchical structure while providing inherent uncertainty quantification. Prior approaches predominantly rely on the Poincaré ball model, which suffers from numerical instability, optimization, and computational challenges. 
We propose a novel, tractable, architecture-agnostic semantic segmentation framework (pixel-wise and mask classification) in the hyperbolic Lorentz model. We employ text embeddings with semantic and visual cues to guide hierarchical pixel-level representations in Lorentz space.
This enables stable and efficient optimization without requiring a Riemannian optimizer, and easily integrates with existing Euclidean architectures. 
Beyond segmentation, our approach yields free uncertainty estimation, confidence map, boundary delineation, hierarchical and text-based retrieval,  and zero-shot performance, reaching generalized flatter minima. 
We introduce a novel uncertainty and confidence indicator in Lorentz cone embeddings. Further, we provide analytical and empirical insights into Lorentz optimization via gradient analysis. 
Extensive experiments on ADE20K, COCO-Stuff-164k, Pascal-VOC, and Cityscapes, utilizing state-of-the-art per-pixel classification models (DeepLabV3 and SegFormer) and mask classification models (mask2former and maskformer), validate the effectiveness and generality of our approach. Our results demonstrate the potential of hyperbolic Lorentz embeddings for robust and uncertainty-aware semantic segmentation. Code is available at \url{https://github.com/mxahan/Lorentz_semantic_segmentation}.
\end{abstract}


   




\section{Introduction}
\label{sec:intro}


Semantic segmentation, which assigns a semantic label to every pixel in an image, is a fundamental task in computer vision. State-of-the-art architectures formulate this problem either as per-pixel classification \cite{chen2018encoder}, or as mask classification, where a set of binary masks is predicted and each mask is associated with a single class label \cite{cheng2021per, cheng2022masked}. Despite these structural differences, both paradigms typically treat semantic segmentation as a flat classification task, relying on one-hot encoded labels and modeling categories as mutually exclusive and independent \cite{xie2021segformer}. This formulation overlooks the rich semantic relationships and hierarchical structures inherent in object categories—for instance, “vehicle” as a parent concept of “car,” “bus,” and “truck.” By disregarding such dependencies, existing approaches fail to incorporate meaningful inductive biases, thereby limiting their generalization capability.

Recent advances have attempted to overcome this limitation by integrating visual and textual information through Vision–Language Models (VLMs) such as CLIP \cite{radford2021learning}. Several studies extend this paradigm to semantic segmentation via text-guided supervision, including mask-level vision–language alignment \cite{das2024mta}, pixel–text score maps \cite{rao2022denseclip}, and text-supervised segmentation frameworks \cite{wu2024image,yun2023ifseg}, resulting in improved zero-shot and open-vocabulary performance. Nevertheless, these approaches, along with most contemporary methods, embed both image and text representations in Euclidean space. Such geometry is inherently limited in modeling hierarchical structures \cite{desai2023hyperbolic}, as it treats all points uniformly and fails to efficiently capture the exponential growth of concept subtypes characteristic of semantic taxonomies \cite{nickel2017poincare, ganea2018hyperbolic}.

Hyperbolic spaces provide a geometrically principled alternative, as their negative curvature and exponential volume growth constitute a continuous analogue of tree-like structures \cite{ganea2018hyperbolic, nickel2018learning}. This property enables efficient representation of hierarchical relationships: general parent concepts (e.g., “animal”) can be positioned near the origin, while increasingly specific child concepts (e.g., “dog,” “cat”) are embedded farther away, naturally encoding \texttt{is-a} relations and yielding interpretable, semantically consistent embeddings \cite{nickel2017poincare, desai2023hyperbolic, ibrahimi2024intriguing, nickel2018learning}. Despite their promise, existing hyperbolic approaches in computer vision \cite{desai2023hyperbolic, ibrahimi2024intriguing, khrulkov2020hyperbolic} predominantly focus on global, image-level representations. Extending hyperbolic modeling to dense prediction tasks remains non-trivial: per-pixel classification requires computationally intensive manifold operations at scale, and mask-classification architectures introduce additional structural complexities that further complicate integration within hyperbolic spaces.

Pioneering studies \cite{atigh2022hyperbolic, chen2023hyperbolic, weber2024flattening} have investigated hyperbolic semantic segmentation using the Poincaré ball model, primarily within per-pixel classification frameworks. However, these approaches are prone to numerical instability and gradient approximation errors \cite{nickel2018learning}. The Lorentz (hyperboloid) model in Minkowski space provides a mathematically equivalent yet more numerically stable and computationally efficient alternative, particularly in low-dimensional settings, offering closed-form expressions for key operations such as distance computation and exponential maps \cite{nickel2018learning}. Despite these advantages and its suitability for large-scale, text-guided segmentation, the Lorentz model remains largely underexplored in dense prediction tasks. Existing Lorentz-based methods predominantly focus on embedding entire images or global features, leaving unresolved the challenge of learning efficient, pixel-wise hyperbolic representations guided by semantic structure. Moreover, current hyperbolic segmentation approaches are limited to per-pixel classification models, due to the relative simplicity of replacing the final Euclidean classification layer with hyperbolic operations. In contrast, mask-classification architectures, despite consistently outperforming per-pixel counterparts, have been largely overlooked in hyperbolic settings. This reveals a clear research gap: the absence of a unified, efficient, and stable hyperbolic framework that systematically supports both pixel-classification and mask-classification paradigms for semantic segmentation.

To address this gap, \cite{hasan2026lorentz} develops a hyperbolic Lorentz framework for per-pixel segmentation architectures. Building upon this foundation, the present work proposes a generalized text-guided hyperbolic semantic segmentation framework in the Lorentz model that supports both per-pixel and mask-classification architectures. We leverage rich textual descriptions to encode semantic and visual priors as class prototypes embedded in hyperbolic space, grounded in the insight that image pixels represent specific instances of abstract concepts described by language, forming an inherent semantic hierarchy. We capture this structure by leveraging the order embeddings~\cite{vendrov2015order} within the Lorentz manifold and extend the entailment loss formulation~\cite{ganea2018hyperbolic, desai2023hyperbolic} to the dense prediction setting. This constrains each pixel-level feature to lie within the convex entailment cone induced by its corresponding textual embedding, enforcing a geometrically consistent visual–semantic hierarchy and enabling structured representation learning for segmentation.

Our main contributions are as follows.
(i)
We propose a \textbf{Novel Text-Guided Hierarchical Hyperbolic Semantic Segmentation Framework} grounded in the numerically stable Lorentz model, enabling scalable pixel-wise training on large datasets while preserving interpretability via Poincaré projection. 
We extend the entailment loss to pixel-level embeddings, constraining visual features within the entailment cone of their corresponding concept, to maintain semantic hierarchy and consistency in hyperbolic space. 
We further provide an analytic gradient analysis to characterize the learning dynamics within the Lorentz model and offer theoretical insight into its optimization behavior.
(ii) Our architecture-agnostic approach introduces \textbf{Minimal Overhead and Architecture Compatibility Across} both per-pixel and mask-classification segmentation models in the Lorentz model.
For per-pixel classification, we adopt a hyperbolic sample-to-prototype formulation for per-pixel architectures. 
For mask-classification architectures, we propose a hyperbolic sample-to-prototype formulation to align mask queries with class prototypes, and a sample-to-sample formulation for semantic mask generation. 
(iii) \textbf{Empirical Validation and Geometric Analysis.}
We validate our approach on multiple benchmarks, including ADE20K, COCO-Stuff-164k, Cityscapes, and Pascal VOC, demonstrating state-of-the-art performance across both per-pixel and mask-classification models. Furthermore, the Lorentz-space geometry provides a novel pixel-level uncertainty estimation by quantifying deviations from euclidean prototype setting, enabling confidence assessment and boundary localization without additional training or post-processing. Extensive ablation studies confirm the effectiveness and robustness of the proposed framework.

\section{Related Works}

\textbf{Semantic Segmentation Architectures}: 
Semantic segmentation architectures generally fall into two paradigms: per-pixel classification and mask classification. Per-pixel approaches assign a class label to each pixel by applying a classification loss at every spatial location, thereby implicitly partitioning the image into semantic regions. Representative models in this paradigm include U-Net \cite{ronneberger2015u}, DeepLabV3 \cite{chen2017rethinking}, and SegFormer \cite{xie2021segformer}. In contrast, mask-classification models predict a finite set of binary masks, each paired with a class probability. This formulation provides a unified framework for semantic, instance, and panoptic segmentation, and has been adopted by architectures such as Mask R-CNN \cite{he2017mask}, DETR \cite{carion2020end}, MaskFormer \cite{cheng2021per}, and Mask2Former \cite{cheng2022masked}.

\textbf{Deep Learning in Hyperbolic Space} has recently garnered significant attention for its capacity to model hierarchical structures and uncertainty, as well as to improve representational robustness \cite{mettes2024hyperbolic, peng2021hyperbolic}. Foundational contributions include Poincaré embeddings for hierarchical representations \cite{nickel2017poincare} and Lorentz model optimization for improved numerical stability and scalability \cite{nickel2018learning}. Comprehensive surveys \cite{mettes2024hyperbolic, peng2021hyperbolic} highlight the advantages of hyperbolic geometry for constructing low-dimensional, hierarchy-preserving embeddings, and document its recent advances in deep learning adaptation. Order embeddings \cite{vendrov2015order} align visual–semantic hierarchies with partial orders in Euclidean space. This framework was subsequently extended to hyperbolic geometry through the introduction of geodesically convex entailment cones in the Poincaré model \cite{ganea2018hyperbolic}. Later works \cite{le2019inferring, desai2023hyperbolic} further adapted this formulation to the Lorentz model, deriving closed-form entailment cone expressions. Complementary to these efforts, \cite{sala2018representation, sarkar2011low} propose algorithms for embedding graph nodes in hyperbolic space with provably low distortion given an underlying tree structure.

\textbf{Text-Guided Semantic Segmentation.}
Open-vocabulary scene parsing aligns image features with textual concepts to predict hierarchical labels \cite{zhao2017open}. DenseCLIP \cite{rao2022denseclip} introduced pixel-text alignment for dense prediction tasks, while PPL \cite{kwon2023probabilistic} incorporated probabilistic prompts to capture diverse semantic attributes. Further approaches leverage mask-level vision-language alignment \cite{das2024mta}, pixel-text score maps \cite{rao2022denseclip}, and text-supervised segmentation frameworks \cite{wu2024image, yun2023ifseg}. CoDe \cite{wu2024image} decomposes paired image-text inputs into constituent regions and segments, enforcing region-word correspondence via contrastive learning with prompt mechanisms. S-Seg \cite{lai2025exploring} proposes leveraging pretrained models to generate pseudo-masks for supervising MaskFormer mask predictions, alongside a text encoder for query-class supervision. SimSeg \cite{yi2023simple} enables fine-tuning of CLIP for semantic segmentation by incorporating text-image locality alignment.

\textbf{Hyperbolic Models for Segmentation.}
Existing hyperbolic space segmentation frameworks primarily integrate the Poincaré ball model within per-pixel classification architectures. Early work \cite{weng2021unsupervised} applied Poincaré embeddings to instance segmentation, while \cite{atigh2022hyperbolic, weber2024flattening} leveraged hypernym–hyponym relations from WordNet \cite{miller1995wordnet} for hyperbolic classification using Poincaré multinomial logistic regression (MLR). HyperUL \cite{chen2023hyperbolic} introduced a hyperbolic uncertainty loss to highlight ambiguous pixels based on hyperbolic distances, and HALO \cite{franco2023hyperbolic} applied hyperbolic neural networks to active learning with a radius-based acquisition score. TOPICS \cite{hindel2024taxonomy} incorporated taxonomy-based hyperbolic regularization for incremental segmentation. Recent studies have developed hyperbolic convolution operations in U-Net decoders \cite{mishra2026hyperbolic} and transformer-based architectures that employ hyperbolic embeddings for pixel-level classification for medical image segmentation \cite{wang2023vison}.

The \cite{yang2024hyperbolic} analyzes the differences between Euclidean and hyperbolic LoRA fine-tuning, finding that the hyperbolic formulation introduces a curvature-dependent length scaling in the parameter update that is better suited for hierarchical representation learning. HyperCLIP \cite{peng2025understanding} extends CAT-Seg \cite{cho2024cat} by introducing a learnable scaling factor in hyperbolic space to improve vision-language alignment. From a theoretical perspective, \cite{mishne2023numerical} establishes the superior numerical stability of the Lorentz model through rigorous gradient analysis. Furthermore, \cite{law2019lorentzian} demonstrates that the Lorentz exponential map preserves ordering with respect to the Euclidean norm.

The most closely related hyperbolic segmentation works are \cite{atigh2022hyperbolic} and \cite{desai2023hyperbolic}. \cite{atigh2022hyperbolic} adopts a \textit{sample-to-gyroplane} strategy, relying on predefined WordNet hierarchies in Poincaré space with computationally expensive Möbius operations, while \cite{desai2023hyperbolic} employs a \textit{sample-to-sample} strategy operating on global image-level representations via contrastive learning in Lorentz space.

In particular, this work extends~\cite{hasan2026lorentz} by broadening hyperbolic semantic segmentation beyond per-pixel classification to mask-classification architectures, introducing an efficient integration of the hyperbolic framework into MaskFormer-style models — a direction unexplored by prior work. Text-guided supervision is further incorporated into both architecture families, addressing the limited exploration of textual guidance in MaskFormer-style models~\cite{lai2025exploring}. Finally, we provide a theoretical analysis of the joint optimization of hyperbolic distance and entailment loss~\cite {desai2023hyperbolic}, offering gradient-level insights into the advantages of the proposed formulation over its Euclidean counterpart.

\section{Preliminaries}

\paragraph{Notation.} We adopt the notation of \cite{nickel2018learning, desai2023hyperbolic}, $\langle \cdot, \cdot \rangle_{\mathcal{L}}$ and $\|\cdot\|_{\mathcal{L}}$ denote the Lorentzian inner product and norm, while $\langle \cdot, \cdot \rangle$ and $\|\cdot\|$ refer to their Euclidean operations.

\subsection{Background: Lorentz (Hyperboloid) Model}

\paragraph{Hyperbolic Space.}  
Hyperbolic space is a unique, complete, and connected $n$-dimensional Riemannian manifold with constant negative curvature \cite{cannon1997hyperbolic}. They cannot be isometrically embedded in Euclidean space $\mathbb{R}^n$. Hyperbolic space can be modeled by several isometric models, of which five are well studied \cite{peng2021hyperbolic, mettes2024hyperbolic}. Among them, the Poincaré and  Lorentz (hyperboloid) models are the most widely used in practice in Deep Learning \cite{nickel2017poincare, nickel2018learning}.

\paragraph{Lorentz Model.}  
The $n$-dimensional Lorentz (hyperboloid) $\mathcal{L}^n = (\mathcal{H}^n, g_{\mathcal{L}})$ model with curvature $-c$ ($c>0$) is defined as the Riemannian manifold in $(n+1)$-dimensional Minkowski space $\mathbb{R}^{1,n}$ ($\mathbb{R}^{n+1}$ equipped with Lorentzian/Minkowski inner product) by
\begin{equation}
    \mathcal{H}^n = \{ \bm{x} \in \mathbb{R}^{1, n} : \langle \bm{x}, \bm{x} \rangle_{\mathcal{L}} = -\frac{1}{c}\}, c> 0
\end{equation}
For $\bm{x}, \bm{y} \in \mathbb{R}^{n+1}$, the Lorentzian inner product is
\begin{equation}
    \langle \bm{x}, \bm{y} \rangle_{\mathcal{L}} = -x_0y_0 + \sum_{i=1}^{n} x_i y_i,
\end{equation}
with norm $
    \|\bm{x}\|_{\mathcal{L}} = \sqrt{|\langle \bm{x}, \bm{x} \rangle_{\mathcal{L}}|}$ and for any point $\bm{x} = (x_0, \bm{x}') \in \mathbb{R}^{1,n}$ is $\bm x \in \mathcal{H}^n$ iff
\begin{equation}
   x_0 = \sqrt{\tfrac{1}{c} + \|\bm{x}'\|^2}.
\end{equation}

\paragraph{Geodesics.}  
Geodesics in $\mathcal{L}^n$ are intersections of the hyperboloid with hyperplanes through the origin, representing the shortest path between two points. The geodesic distance between $\bm{x}, \bm{y} \in \mathcal{L}^n$ is
\begin{equation}
    d_{\mathcal{L}}(\bm{x}, \bm{y}) = \tfrac{1}{\sqrt{c}} \cosh^{-1}\!\big(-c \langle \bm{x}, \bm{y} \rangle_{\mathcal{L}}\big).
    \label{eq:lorentz_dist}
\end{equation}

\paragraph{Tangent Space.}  
The tangent space at $z \in \mathcal{L}^n$ is a linear subspace of $R^{1,n}$ with positive-definite metric ($\langle \bm v, \bm v\rangle_\mathcal{L} > $0 for  $\bm v \neq \bm 0$)
\begin{equation}
    \mathcal{T}_z\mathcal{L}^n = \{\bm{v} \in \mathbb{R}^{1,n} : \langle \bm{v}, \bm{z} \rangle_{\mathcal{L}} = 0\}
\end{equation}
Any ambient $\bm{u} \in \mathbb{R}^{n+1}$ can be projected onto $\mathcal{T}_z\mathcal{L}^n$ by orthogonal projection in terms of the Lorentz product. 
\begin{equation}
   \bm v = \text{proj}_{\bm{z}}(\bm{u}) =  \bm{u} - \bm{z} \frac{\langle \bm{z}, \bm{u} \rangle_{\mathcal{L}}}{\langle \bm z, \bm z \rangle_{\mathcal{L}}}= \bm{u} + c \bm{z} \langle \bm{z}, \bm{u} \rangle_{\mathcal{L}}.
\end{equation}

\paragraph{Exponential and Logarithmic Maps.}  
The exponential map  $\text{expm}_{\bm{z}} : \mathcal{T}_z\mathcal{L}^n \to \mathcal{L}^n$ for vector $\bm v \in \mathcal{T}_{\bm z}\mathcal{L}^n$ 
\begin{equation}
 \bm x =   \text{expm}_{\bm{z}}(\bm{v}) = \cosh(\sqrt{c}\|\bm{v}\|_{\mathcal{L}})\bm{z} + \frac{\sinh(\sqrt{c}\|\bm{v}\|_{\mathcal{L}})}{\sqrt{c}\|\bm{v}\|_{\mathcal{L}}}\bm{v}
 \label{eq:exp_map}
\end{equation}
The inverse logarithmic map $\text{logm}_{\bm{z}} : \mathcal{L}^n \to \mathcal{T}_z\mathcal{L}^n$ is
\begin{equation}
   \bm v = \text{logm}_{\bm{z}}(\bm{x}) = \frac{\cosh^{-1}(-c \langle \bm{z}, \bm{x}\rangle_{\mathcal{L}})}{\sqrt{(c\langle \bm{z}, \bm{x}\rangle_{\mathcal{L}})^2 - 1}} \, \text{proj}_{\bm{z}}(\bm{x}).
\end{equation}

\paragraph{Equivalence of the Models.}
Several isometric and equivalent models of hyperbolic space are widely used, with the Lorentz (hyperboloid), Poincaré disk, and Klein models being the most common. Each has practical advantages: the Poincaré disk offers intuitive visualization and uncertainty interpretation \cite{atigh2022hyperbolic}, the Klein model provides a simple formulation of the Einstein midpoint for averaging, and the Lorentz model is well-suited for optimization and numerical stability. Owing to their equivalence, one can leverage their respective strengths via smooth isometric mappings:

$\mathcal{H}^n \leftrightarrow \mathcal{P}^n$:
\begin{equation}
\bm{p} = \frac{\bm{x}'}{x_0\sqrt{c} + 1}, \quad
x_0 = \frac{1 + c\|\bm{p}\|^2}{\sqrt{c}(1 - c\|\bm{p}\|^2)}, \quad
\bm{x}' = \frac{2\bm{p}}{1 - c\|\bm{p}\|^2}.
\label{eq:l2p}
\end{equation}

$\mathcal{H}^n \leftrightarrow \mathcal{K}^n$:
\begin{equation}
\bm{k} = \frac{\bm{x}'}{x_0\sqrt{c}}, \quad
x_0 = \frac{1}{\sqrt{c(1 - c\|\bm{k}\|^2)}}, \quad
\bm{x}' = x_0 \bm{k}\sqrt{c}.
\label{eq:l2k}
\end{equation}

$\mathcal{K}^n \leftrightarrow \mathcal{P}^n$:
\begin{equation}
\bm{p} = \frac{\bm{k}}{1 + \sqrt{1 - c\|\bm{k}\|^2}}, \quad
\bm{k} = \frac{2 \bm{p}}{1 + c\|\bm{p}\|^2}.
\end{equation}

\paragraph{Hyperbolic Averaging.}
In the Klein model, averaging takes the form of the Einstein midpoint:
\begin{equation}
\text{HypAvg}(\bm{k}_1, \dots, \bm{k}_N) = \frac{\sum_i \gamma_i \bm{k}_i}{\sum_i \gamma_i},
\quad \gamma_i = \frac{1}{\sqrt{1 - c\|\bm{k}_i\|^2}}.
\label{eq:einsten_mp}
\end{equation}

\medskip
In this work, we primarily work with the Lorentz model in Minkowski space, converting to alternative representations when beneficial. We retrieve the notation of MERU \cite{desai2023hyperbolic} by setting the temporal component $x_0 = x_{time}$ and the spatial component $\bm{x}' = \bm{x}_{space} \in \mathbb{R}^n$.

\subsection{Metric Tree-likeness: Gromov $\delta$-Hyperbolicity}


We evaluate how closely the semantic embeddings of pixels follow a tree-like structure for four datasets. For a Gromov-hyperbolic metric space, there exists a $\delta \geq 0$ such that every geodesic triangle is $\delta$-slim \cite[Def.~1.16]{bridson2013metric}. Geodesic slimness captures negative curvature, and $\delta$-hyperbolicity characterizes how tree-like a metric space is in terms of its metric structure \cite{gromov1987hyperbolic,adcock2013tree} and is used to assess the hyperbolicity of datasets and embeddings \cite{peng2021hyperbolic}.

We adopt the $\delta$-hyperbolic metric proposed in \cite{khrulkov2020hyperbolic} to assess semantic dataset hyperbolicity. 
The Gromov product is defined for $y, z, r \textit{ $($base point$)$} \in \bm X$ as $ (y|z)_r = \tfrac{1}{2}\big(d(r, y) + d(r, z) - d(z, y)\big)$. Then, $\delta$ is the minimal value such that the following four-point condition holds for all $x, y, z, r \in \bm X$:
\begin{equation}
    (x|z)_r \geq \min \{(x|y)_r, (y|z)_r \} - \delta.
\end{equation}

To compute $\delta$, we employ the efficient fixed base point exact max–min product algorithm (complexity $\mathcal{O}(n^{2.69})$) \cite{fournier2015computing}. For the matrix $A$ of pairwise Gromov products, $\delta = \underset{i,j}{\max} [(A \otimes A)_{ij} - A_{ij}]$, where $(A \otimes B)_{ij} = \underset{k}{\max} \min\{A_{ik}, B_{kj}\}$ \cite{adcock2013tree}. Further, \cite{khrulkov2020hyperbolic} proposed the scale-invariant metric  $\delta_{rel}(X) = \tfrac{2\delta(X)}{diam(X)}\in [0,1] $ 
and $diam(X)$ denotes the maximal pairwise distance. Intuitively, a small $\delta_{rel}$ indicates a space closer to a tree structure (strong hyperbolicity).

\textbf{Experiments and Findings.}  
Due to computational constraints, we adopt batch-wise estimation of the hyperbolicity measure $delta_{rel}$ across both per-pixel and mask-classification models. All models were pretrained under standard Euclidean settings.

\textbf{Per-Pixel classification models.}
We compute $\delta_{rel}$ on pretrained SegFormer and DeepLabV3 encoder embeddings for input images of resolution $(512 \times 512)$ from four benchmark semantic segmentation datasets. The SegFormer encoder produces four feature maps with channel dimensions $(64, 128, 320, 512)$ at spatial resolutions $(128, 64, 32, 16)$, respectively, while the DeepLabV3 encoder (ResNet-101 backbone) yields a single feature map of dimension $2048$ at $64 \times 64$ resolution. For each feature set and dataset, $\delta_{rel}$ is computed over randomly sampled batches and the average is reported in Tab.~\ref{tab:del_test_result}. The obtained $\delta_{rel}$ values are significantly closer to $0$ than to $1$ across all feature sets from all four datasets, indicating strong latent hyperbolicity in both models and suggesting that visual representations naturally lend themselves to hyperbolic embedding. Text embeddings similarly exhibit a tree-like organization, with an average $\delta_{rel} \approx 0.3$ across the four evaluated datasets. Notably, SegFormer features yield relative hyperbolicities of $(0.31, 0.24, 0.18, 0.18)$ at successive encoder resolutions, demonstrating that coarser, lower-resolution features consistently exhibit stronger hyperbolicity — consistent with their encoding of more global semantic structure.

\textbf{Mask classification models.}
We compute $\delta_{rel}$ on pretrained MaskFormer and Mask2Former (base and large variants) pixel-level module encoder and decoder embeddings for input images of resolution $(640 \times 640)$ from the two largest semantic segmentation datasets. For large and base variants of both models, the encoder's last hidden state produces feature maps at a spatial resolution of $(20 \times 20)$, with feature dimensions of $1536$ and $1024$, respectively. The decoder's last hidden state produces feature maps of dimension $256$ at a spatial resolution of $(160 \times 160)$ for both variants. Results are reported in Tab.~\ref{tab:del_test_result_mf}. Consistent with the per-pixel models, maskformer image embeddings exhibit strong hyperbolicity, with  $\delta_{rel}$ values closer to $0$ for both datasets. Furthermore, encoder embeddings demonstrate a stronger tree-like structure than decoder embeddings across both datasets, attributable to their encoding of higher-level semantic label features. Additionally, large model variants exhibit marginally stronger hyperbolicity than their base counterparts.

\begin{table}[h!]
\centering
\footnotesize
\resizebox{\columnwidth}{!}{
\begin{tabular}{l|cccc}
\hline
\textbf{Encoders} & Pascal-VOC~\cite{everingham2010pascal} & Cityscapes~\cite{cordts2016cityscapes} & ADE20K~\cite{zhou2019semantic} & COCO-Stuff~\cite{caesar2018coco} \\
\hline
DeepLabV3~\cite{chen2017rethinking}                & 0.23 & 0.24      & 0.26   & 0.23  \\
SegFormer~\cite{xie2021segformer}                & 0.24 & 0.23      & 0.23   & 0.22  \\
\hline
\end{tabular}
}
\caption{The relative delta $\delta_{rel}$ values for different semantic datasets. We compute the Euclidean distance between the encoder features and report the average over 32 randomly sampled batches of size 1024. Smaller values of $\delta_{\mathrm{rel}}$ indicate stronger hyperbolicity}
\label{tab:del_test_result}
\end{table}

\begin{table}[t]
\centering
\footnotesize
\begin{tabular}{p{1.6cm}  p{1 cm} p{1cm} p{1.1 cm} p{1.1 cm}}
\toprule
Model (Size) & Module & Training Data & Tested on COCO & Tested on ADE \\
\midrule
MaskFormer  & Encoder & COCO & 0.28 & 0.27 \\
(Large)    & Decoder & COCO & 0.38 & 0.33 \\
                   & Encoder & ADE  & 0.31 & 0.33 \\
                   & Decoder & ADE  & 0.40 & 0.39 \\
\midrule
MaskFormer & Encoder & COCO & 0.32 & 0.29 \\
 (Base) & Decoder & COCO & 0.33 & 0.33 \\
                & Encoder & ADE  & 0.28 & 0.31 \\
                    & Decoder & ADE  & 0.34 & 0.40 \\
\midrule
Mask2Former  & Encoder & COCO &  0.26  &  0.32 \\
(Large)    & Decoder & COCO &  0.34  & 0.38 \\
                   & Encoder & ADE  & 0.32 & 0.31\\
                   & Decoder & ADE  & 0.38  & 0.36 \\
\bottomrule
\end{tabular}
\caption{The relative delta $\delta_{rel}$ values for different semantic datasets for trained maskformer models. Smaller values of $\delta_{\mathrm{rel}}$ indicate stronger hyperbolicity}
\label{tab:del_test_result_mf}
\end{table}

\section{Methodology} \label{sec:method}
\begin{figure}
    \centering
    \includegraphics[width = 1 \linewidth]{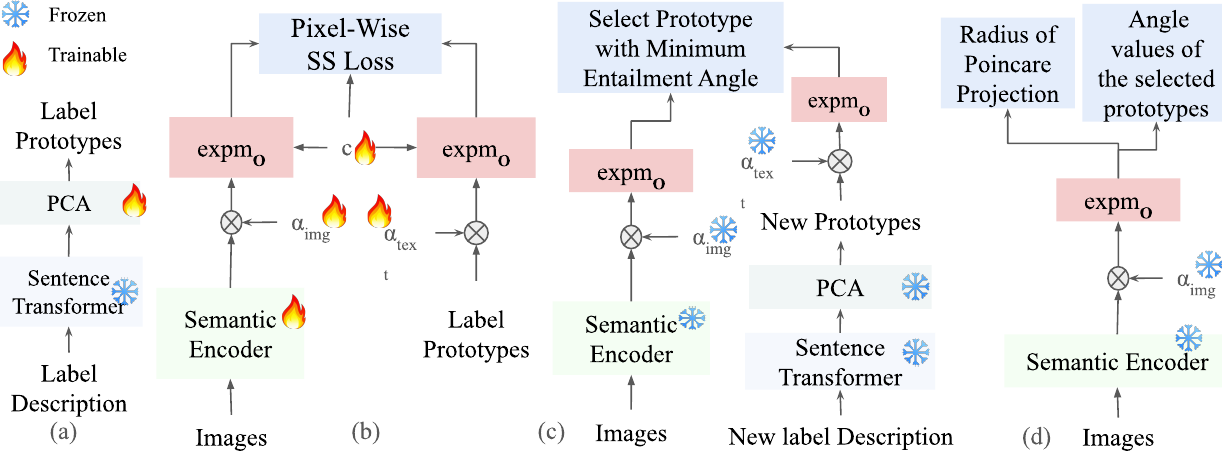}
    \caption{(a) Label Information encoding (b) Lorentz Space Optimization for the semantic segmentation (c) Zero-shot pixel label identification (d) Model Uncertainty Estimation}
    \label{fig:overview}
\end{figure}

Our framework enables Lorentz hyperbolic semantic segmentation for both \textit{per-pixel classification} and \textit{mask-classification} architectures. In per-pixel classification models, the network produces an embedding for each pixel and predicts a probability distribution over $C$ semantic categories. Extending this formulation to the hyperbolic setting is straightforward: pixel embeddings are aligned with textual label embeddings within the Lorentz manifold, allowing per-pixel semantic supervision in hyperbolic space.

In contrast, Mask classification architectures take a fundamentally different approach, grouping pixels into $N$ segments by predicting $N$ binary masks alongside their corresponding category labels, enabling flexible handling of segmentation tasks. The meta-architecture consists of three components: a backbone encoder, a pixel decoder, and a transformer decoder. The backbone extracts multi-scale visual features, which are refined by the pixel decoder and consumed by the transformer decoder to produce $N$ mask-feature pairs that are combined to generate the final segmentation output.

While both paradigms share common objectives, their architectural differences require distinct hyperbolic formulations. Our approach aligns textual label representations and pixel-level visual features in a shared Lorentz embedding space, analogous to vision–language alignment. Specifically, we extend the Lorentz model to support efficient dense computations and employ entailment-cone constraints to guide pixel-level supervision.

\textbf{Label-Description Pre-processing (Fig \ref{fig:overview}(a)).} For each of the $C$ semantic classes, we define a descriptive textual representation capturing both semantic and visual attributes. 
These descriptions can be manually curated (e.g., \textit{bus}: public transport buses, rectangular shape, windows and doors visible, often on roads, \textit{Frisbee:} Flat, round plastic disc used in games; brightly colored and flies through the air when thrown.) or sourced automatically from WordNet \cite{miller1995wordnet} definitions. We encode each textual description using Sentence Transformers \cite{reimers2019sentence}, producing a $d_\text{orig} = 384$ dimensional embedding. We perform PCA on the label features and reduce the dimension d to get the encoded class prototype $\bm{x}_{enc, i}$ for the $i$-th class. 
The resulting set of $C$ prototypes serves as semantic anchors in the embedding space for both supervised alignment and entailment cone construction. 
Tab. \ref{tab:sentence_encoder} shows the label similarity statistics for the resulting embeddings.

\begin{table}[]
   
    \centering
     \footnotesize
    \begin{tabular}{l|c}
    \hline
    Label & Relevant Classes (similarity Index) \\ \hline
    Dog   & cow (0.77), cat (0.74), horse (0.66), sheep (0.63)    \\
    Chair & couch (0.73), table (0.63), bed (0.58), 'bench (0.57) \\ \hline
    \end{tabular}
    \caption{Semantic similarity between class descriptions: Top-4 nearest classes identified using cosine similarity scores (COCO-Stuff).}
    \label{tab:sentence_encoder}
\end{table}

\paragraph{Image Embedding.}
\textbf{Per-pixel Model: }
Input images of shape $[h, w, 3]$ are passed through a semantic segmentation network, yielding dense feature maps of shape $[h, w, d]$, where $d$ matches the dimension of the text embeddings. For each pixel $j$, the corresponding feature vector is denoted $\bm{y}_{\text{enc}, j}$.

\textbf{Mask Classification Models:}
The mask-classification processes the input image through three sequential components. First, a backbone encoder extracts multi-scale visual features, producing low-resolution feature maps at multiple output resolutions. Second, a pixel decoder progressively upsamples these low-resolution features to generate high-resolution per-pixel embeddings of shape $[h, w, d_\epsilon]$. Third, a transformer decoder employs $N$ learnable queries to cross-attend to the lowest-resolution feature maps, aggregating global information about each segment. This produces $N$ mask feature vectors of shape $[N, d_\epsilon]$. Finally, binary mask predictions are decoded by combining the per-pixel embeddings with the $N$ mask feature vectors via dot-product attention, yielding $N$ segment masks.

\subsection{Semantic Segmentation in Lorentz Space}

Following \cite{desai2023hyperbolic}, we lift both text and image embeddings to Lorentz hyperbolic space $\mathcal{H}^n$. Unlike prior work that used a single image embedding, semantic segmentation requires per-pixel embeddings, which poses computational challenges. We adopt strategies from \cite{atigh2022hyperbolic} and devise efficient computational steps for the hyperbolic Lorentz model, enabling scalable pixelwise hyperbolic operations across both per-pixel and mask-classification architectures.

\subsection{Lifting to Lorentz Model}

Both text and image embeddings initially reside in ambient Euclidean space. To perform hyperbolic operations, we lift these embeddings into the Lorentz model::

\textbf{Projection on Lorentz.}
Let the ambient vector be $\bm u = [0, \bm v_e]$ ($\bm v_e \in \{\bm x_{\text{enc}}, \bm y_{\text{enc}}\}$). The Lorentz origin $\bm O = [\frac{1}{\sqrt{c}}, \bm 0] \in \mathbb{R}^{n+1}$ with $\bm 0 = [0, ..., 0] \in \mathbb{R}^n$. Similar to \cite{desai2023hyperbolic},  we consider the origin of the hyperboloid for the projection. Projection to the tangent space $\mathcal{T}_{\bm O}\mathcal{L}^n$ is trivial since $\langle \bm u, \bm O\rangle_\mathcal{L} = 0$, i.e., $\bm v = \text{proj}_{\bm O}(\bm u) = \bm u$.

\textit{Exponential Map:}
$\bm v$ is lifted to the Lorentz model, $\bm h = \text{expm}_{\bm O}(\bm v)\in \mathcal{H}^n$, using the exponential map of Eq. \ref{eq:exp_map} :
\begin{equation}
\bm h = [h_0, \bm h'] = \Big[\frac{\cosh(\sqrt{c} \|\bm v\|_\mathcal{L})}{\sqrt{c}}, \frac{\sinh(\sqrt{c} \|\bm v\|_\mathcal{L})}{\sqrt{c} \|\bm v\|_\mathcal{L}} \bm v_e\Big] ,
\label{eq:l_map_O}
\end{equation}
For memory efficiency, we precompute $\|\bm v_e\|$.

\subsection{Hyperbolic Similarity}

\begin{figure}
    \centering
    \includegraphics[width=0.7\linewidth]{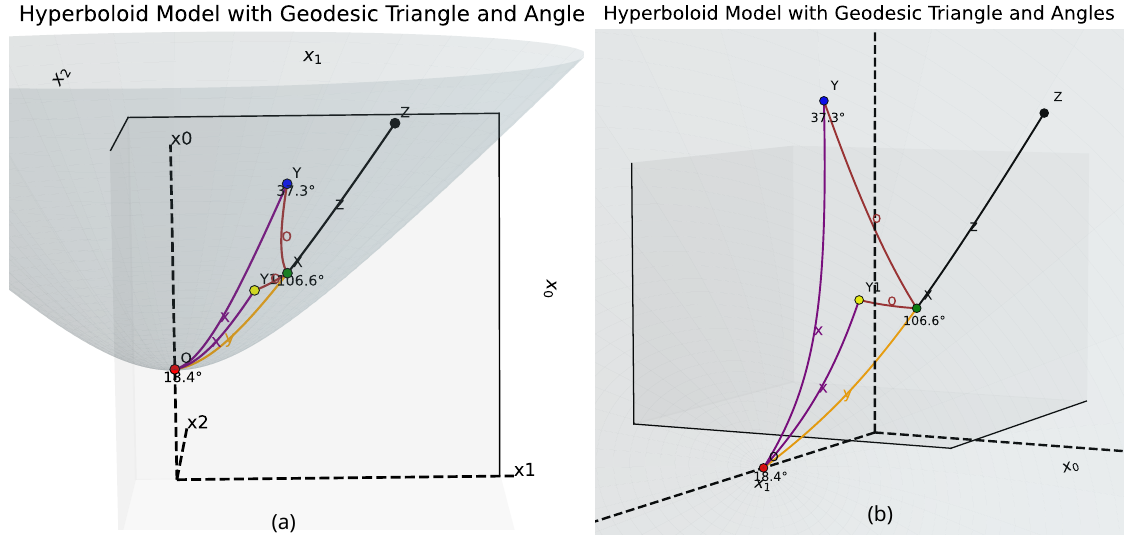}
    \caption{Hyperbolic triangle for Y and Y1 points for pivot X in Lorentz surface. Despite Y1 being closer to X, it has higher entailment loss due to its direction w.r.t to X. (a) Full, (b) Zoomed view.}
    \label{fig:loss_ent}
\end{figure}

We consider two complementary similarity measures in the Lorentz model for a given anchor embedding: geodesic distance and entailment cone angle. Together, these capture both proximity and hierarchical order in hyperbolic space.

\textbf{Lorentz Geodesic Distance.}
For two embeddings in the Lorentz space $\rm x_i \in \mathcal{H}^n$ and $\rm y_j  \in \mathcal{H}^n $, their proximity can be measured using the Lorentz geodesic distance:
\begin{equation}
d_\mathcal{L}(\bm x_i, \bm y_j) = \frac{1}{\sqrt{c}} \cosh^{-1}(-c \langle \bm x_i, \bm y_j \rangle_\mathcal{L}),
\end{equation}
While negative of geodesic distance captures similarity, it does not encode hierarchical ordering between embeddings \ref{fig:loss_geom_2d}.

\textbf{Aperture Angle as Entailment Cone.}
To model hierarchical relations, we employ entailment cones that enforce order constraints in hyperbolic space. Convex entailment cones on Riemannian manifolds define a partial order satisfying axial symmetry, rotational invariance, continuous aperture control, and transitivity \cite{ganea2018hyperbolic}. The half-aperture angle of the cone for anchor $\rm x_i$ is defined as
\begin{equation}
\text{aper}(\bm x_i) = \sin^{-1} \Big(\frac{2K}{\sqrt{c} \, \|\bm x_i'\|} \Big),
\label{eq:aperture}
\end{equation}
where $\| \rm x_i'\|$ denotes the norm of the spatial component of the Lorentz embedding and $K$ is a constant.

\textbf{Lorentz Exterior Angle as Entailment Similarity.}
For a hyperbolic triangle formed by the origin $\mathbf{O}$, anchor $\mathbf{x}_i$, and feature embedding $\mathbf{y}_j$, we define the exterior angle with $\mathbf{x}_i$ as the pivot:
\begin{equation}
\text{ext}(\bm x_i, \bm y_j) = \pi - \angle \bm O \bm x_i \bm y_j.
\end{equation}
Ideally, $\angle \mathbf{O}\mathbf{x}_i\mathbf{y}_j \rightarrow \pi$ indicates that the feature embedding lies within the entailment cone of the anchor. Using the hyperbolic law of cosines, \cite{desai2023hyperbolic} derived the following closed-form expression:
\begin{equation}
\text{ext}(\bm x_i, \bm y_j) = \cos^{-1} \Bigg( \frac{y_{0,j} + x_{0,i} \, c \langle \bm x_i, \bm y_j \rangle_\mathcal{L}}{\|\bm x_i'\| \, \sqrt{(c \langle \bm x_i, \bm y_j \rangle_\mathcal{L})^2 - 1}} \Bigg).
\label{eq:ext_angle}
\end{equation}
A larger exterior angle indicates weaker alignment and lower semantic similarity between the feature and the anchor.

\subsection{Model Adaptation}

\begin{figure}
    \centering
    \includegraphics[width=1\linewidth]{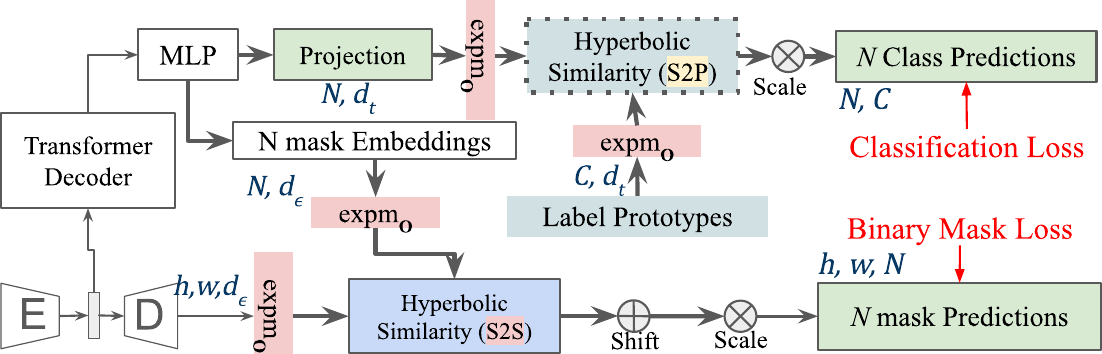}
    \caption{Mask former architecture adaptation for the hyperbolic model. The sample to prototype (s2p) and sample to sample (s2s) blocks are marked. The highlighted blocks show our modification from the original architectures. }
    \label{fig:overview_mf}
\end{figure}

\paragraph{Hyperbolic Per-pixel classification Models}

We adopt a \textit{sample-to-prototype} formulation for the per-pixel model, where each pixel embedding is aligned with the corresponding class prototype in Lorentz space. Let $\bm x_i = \text{expm}_{\bm O}([0, \bm x_{enc, i}])$ denote the Lorentz embedding of the textual prototype for class $i$ among $C$ classes, and $\bm y_j = \text{expm}_{\bm O}([0, \bm y_{enc, j}])$ the Lorentz embedding of pixel $j$. For each pixel, we compute its similarity with all class prototypes using the Lorentz geodesic distance.

\textbf{Cross-Entropy Loss in Lorentz.}
We use pixel-wise logits via the negative Lorentz geodesic distance to each class prototype. The cross entropy loss for the $j$-th pixel with ground-truth class $i$ is
\begin{equation}
l_{ce}(j) = -\log \big( \frac{\exp(-d_\mathcal{L}(\bm x_i, \bm y_j)/\tau)}{\sum_k^{C} \exp(-d_\mathcal{L}(\bm x_k, \bm y_j)/\tau)}\big).
\end{equation}
where $\tau$ is a temperature parameter controlling the distribution sharpness. This loss encourages each pixel embedding to be closer in Lorentz space to its corresponding class prototype than to all other prototypes.

\textbf{Lorentz Entailment Loss.}
The entailment loss enforces a hierarchical geometric constraint: each pixel embedding $\bm{y}_j$ should lie within the entailment cone defined by its corresponding class prototype $\bm{x}_i$, which acts as the conceptual anchor. The half-aperture of the cone at $\bm{x}_i$ is given by Eq.~\ref{eq:aperture}, and the exterior angle $\text{ext}(\bm{x}_i, \bm{y}_j)$ is computed via Eq.~\ref{eq:ext_angle}. The entailment loss penalizes pixel embeddings that fall outside the aperture of their corresponding class cone:
\begin{equation}
\mathcal{L}_{entail} (\bm x_i, \bm y_j) = \max \big(0, \text{ext}(\bm x_i, \bm y_j) - \text{aper}(\bm x_i) \big).
\end{equation}

\textit{Final Loss:}
For overall loss of per-pixel models, we combine cross-entropy and entailment loss components with $\lambda_w$ as a weighting factor, balancing classification accuracy and hierarchical consistency.  
\begin{equation}
\mathcal{L} = \sum_j \big( l_{ce}(j) + \lambda_w \, \mathcal{L}_{entail}(j) \big),
\end{equation}

\paragraph{Hyperbolic Mask classification Models} 
We extend MaskFormer-style architectures (e.g., MaskFormer and Mask2Former) to the Lorentz manifold with minimal architectural changes. 
Mask-classification models predict a set of $N$ binary masks and associate each mask with a semantic category. 
Our formulation introduces two hyperbolic components: 
(i) a \textit{sample-to-prototype} formulation for class prediction, and 
(ii) a \textit{sample-to-sample} formulation for mask generation, as depicted in figure \ref{fig:overview_mf}.

\textbf{Class Query Logits (Sample-to-Prototype).}
In the original (Euclidean) MaskFormer, a linear classifier followed by a softmax predicts class probabilities over $C$ categories for $N$ decoder query embeddings.
We replace this with a hyperbolic sample-to-prototype formulation that enables text-guided supervision.
Let $\bm{x}_i$ denote the Lorentz embedding of the $i$-th class prototype ($i=1,\ldots,C$), and  $\bm{q}_j$ denote the Lorentz embedding of the $j$-th query ($j=1,\ldots,N$) obtained via exponential mapping, and . 
We compute the class-query logits as follows:
\begin{equation}
q_{ij} = 
-w_d\, d_{\mathcal{L}}(\bm x_i,\bm{q}_j)
-
\max\big(0,\text{ext}(\bm x_i, \bm{q}_j)-\text{aper}(\bm x_i)\big),
\end{equation}
producing a logit matrix $\mathbf{Q}\in\mathbb{R}^{N\times C}$. 
The distance term is scaled by $w_d$ to balance its range with the entailment component. 
Softmax over the $C$ classes yields the class probabilities for each query.

\textbf{Mask Query Logits (Sample-to-Sample).}
In the Euclidean formulation, mask logits are obtained via a Euclidean dot product between $N$ mask query embeddings $(N,d_\epsilon)$ and a $ d_\epsilon$-dimensional high-resolution per-pixel feature map of $(h,w)$. 
We replace this with hyperbolic similarity between mask embeddings and pixel features. 
Let $\bm{m}_i$ denote the Lorentz embedding of the $i$-th mask query and $\bm{p}_e$ the Lorentz embedding of the pixel decoder output. 
For each query, we compute distance- and angle-based similarities:
\begin{equation}
\bm m_i^d = \frac{-d_{\mathcal{L}}(\bm{m}_i,\bm{p}_e)+b_d}{s_d},
\qquad
\bm m_i^a = \frac{-\text{ext}(\bm{m}_i,\bm{p}_e)+b_a}{s_a},
\label{eq:scale_shift}
\end{equation}
where $b_d,b_a$ are bias terms that shift the distance and angle values into a suitable range for sigmoid activation, and $s_d,s_a$ are scaling parameters controlling sensitivity. 
The final mask logits for query $i$ are:
\begin{equation}
\bm{m}_i^q = \bm m_i^d + \bm m_i^a , 
\end{equation}
Where each $\bm{m}_i^*\in \mathbb{R}^{h\times w}$. 
Applying a sigmoid function produces the $N$ predicted binary masks for $i = 1, \dots, N $.

\textbf{Loss Optimization.}
The network is trained using the standard mask-classification objective. Predicted masks are first matched to ground-truth annotations via bipartite Hungarian matching to obtain the optimal query-to-annotation assignment. The total loss comprises a cross-entropy term for class prediction and focal and dice loss terms for mask prediction, applied to the matched pairs.

\paragraph{Efficient Computation.}
The prototype distance cross-entropy and entailment losses rely on repeated computations. We precompute per-pixel $x_0$, $\bm x'$, $c \langle \bm x, \bm y \rangle_\mathcal{L}$, and $\|\bm x'\| = \sqrt{x_0^2 - 1/c}$ for memory-efficient hyperbolic operations.

\paragraph{Efficiency over Poincaré.}
Compared to \cite{atigh2022hyperbolic, weber2024flattening}, our Lorentz solution is substantially more efficient. Previous \textbf{sample-to-gyroplane} formulation in the Poincaré ball requires computationally expensive Möbius additions with gyroplane offsets and dot products with gyroplane orientations for pixel-wise class probability estimation \cite{ganea2018hyperbolicNN}. In contrast, our approach in the Lorentz model requires only a single Lorentz inner product—equivalent to a Euclidean dot product with a sign reversal on the time-like component—thereby yielding a major computational advantage.

\subsection{Gradient Analysis}

We analyze the gradient behavior of the Lorentz hyperbolic distance and entailment losses to characterize how hyperbolic operations shape the learning dynamics relative to a purely Euclidean setting.

\textbf{Gradient Interaction.}
We study the interaction between angle- and distance-based losses in the Lorentz model. Let $L=\langle \bm{x}, \bm{y} \rangle_{\mathcal{L}}$. The cosine similarity between their gradients is governed by $\mathrm{sign}((-x_0 L) - y_0)$ (see appendix). The sign determines whether the objectives exhibit \emph{gradient affinity} (positive cosine similarity) or \emph{conflict} (negative cosine similarity) \cite{yu2020gradient,du2018adapting,vu2022spot}.

In a multi-task setting, two loss components are said to exhibit \textit{gradient affinity} when their cosine similarity is positive~\cite{vu2022spot, du2018adapting}, accelerating joint optimization, and \textit{gradient conflict} when their cosine similarity is negative~\cite{yu2020gradient}, slowing convergence. We establish the following result for the joint hyperbolic distance and entailment losses:

\vspace{0.5em}
\noindent\textit{Theorem.} The cosine similarity between the gradients of the distance and entailment loss components is determined by $\mathrm{sign}((-x_0 L) - y_0)$, where $L = \langle \bm{x}, \bm{y} \rangle_{\mathcal{L}}$.
\vspace{0.5em}

When $\mathrm{sign}((-x_0 L) - y_0) > 0$, gradients are aligned (angle $<90^\circ$), leading to synergistic updates and faster convergence. When negative, typically near uncertain regions (e.g., $x_0 < y_0$ with higher $L$), gradients conflict, resulting in smaller updates. In contrast, Euclidean counterparts remain orthogonal, yielding zero cosine similarity and uniform interaction.

This position-dependent interaction induces implicit adaptive gradient update regions: uncertain regions receive conservative updates, while confident regions benefit from accelerated optimization, consistent with prior findings on gradient affinity \cite{li2024scalable}. Our observations align with \cite{franco2023hyperbolicsp}, which reports self-paced behavior in hyperbolic Poincaré embeddings. Similarly, we observe stronger gradients for both loss components near ground-truth regions in the Lorentz model.

\textbf{Gradient Coupling via the Exponential Map.}
Unlike Euclidean space, hyperbolic embeddings introduce globally coupled gradients via the exponential map. For tangent vector $\mathbf{v}$ and Lorentz embedding $\bm x$:
\begin{equation}
\frac{\partial x_j}{\partial v_l} = \frac{\delta_{jl}}{|\mathbf{v}|} \sinh(|\mathbf{v}|) + \frac{v_j v_l}{|\mathbf{v}|^2} \left( \cosh(|\mathbf{v}|) - \frac{\sinh(|\mathbf{v}|)}{|\mathbf{v}|} \right).
\end{equation}
The diagonal term is always positive, while the cross-term depends on $v_j v_l$ as $\cosh(|\mathbf{v}|) - \frac{\sinh(|\mathbf{v}|)}{|\mathbf{v}|} > 0$. Consequently, each tangent space coordinate influences all hyperbolic embedding coordinates, and conversely, the gradient of any hyperbolic coordinate propagates back to all tangent space coordinates. Such dependencies are absent in Euclidean space.

\textbf{Empirical Analysis.}
To isolate and verify these theoretical findings, we conduct simplified gradient experiments in which a target prototype is fixed, and gradients are computed after the exponential map with respect to both the hyperbolic embeddings and the Euclidean tangent space coordinates, under the distance and entailment losses separately. We visualize the resulting gradient directions and magnitudes in 2D to illustrate differences in loss landscapes and gradient fields, confirming: (i) Euclidean gradients of the distance and entailment losses are orthogonal; (ii) hyperbolic gradients exhibit location-dependent relative angles governed by the sign condition above; and (iii) both distance and angle-based losses yield higher gradient magnitudes near ground-truth regions, providing empirical evidence of adaptive, self-paced learning in the Lorentz model.

\begin{figure}
    \centering
    \includegraphics[width=1\linewidth]{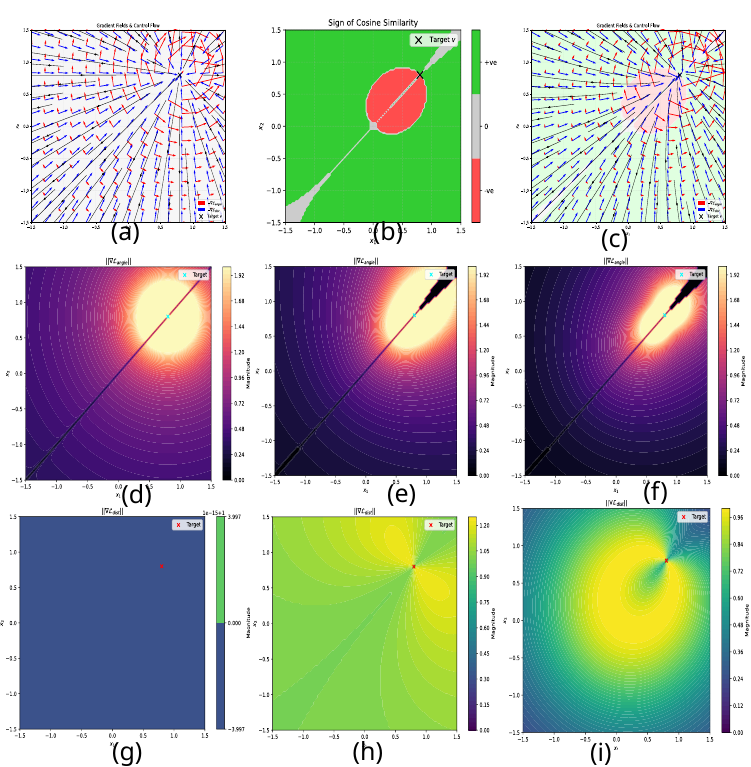}
    \caption{Directions for the (a) Euclidean distance (blue) and Entailment (red), (c) Lorentz distance (blue) and Entailment (red), (b) Sign of the gradients cosine similarity. $||\nabla \mathcal{L}_{angle}||$ and $||\nabla \mathcal{L}_{distance}||$ for Euclidean w.r.t coordinates (d) and (g), hyperbolic w.r.t the tangent coordinates (e) and (h), hyperbolic w.r.t the exponential map coordinates (f) and (i).}
    \label{fig:gradient}
\end{figure}








\section{Experiment and Results}

\textbf{Datasets.}
We evaluate on four standard semantic segmentation benchmark datasets. 
(i) ADE20K~\cite{zhou2019semantic} contains 151 indoor and outdoor categories spanning natural scenes, artifacts, vehicles, food, and animals. 
(ii) COCO-Stuff 164K~\cite{caesar2018coco} extends MS COCO~\cite{lin2014microsoft} with pixel-level annotations for 182 object and “stuff” classes such as sky, grass, and road. 
(iii) Pascal VOC 2012~\cite{everingham2010pascal} includes 20 object categories covering everyday objects, animals, and vehicles. 
(iv) Cityscapes~\cite{cordts2016cityscapes} provides fine-grained annotations for urban street scenes with categories such as road, vehicles, and pedestrians. 
For each dataset, we incorporate visual-semantic textual descriptions for all labels.

\textbf{Per-Pixel Classification Models.}
We evaluate our framework on two SOTA representative per-pixel classification architectures.
\textit{DeepLabV3}~\cite{chen2017rethinking} employs atrous spatial pyramid pooling to capture multi-scale contextual information. It supports several backbone variants; in our experiments, we adopt a ResNet-101 backbone pretrained on Pascal VOC~\cite{everingham2010pascal}. \textit{SegFormer}~\cite{xie2021segformer} is a transformer-based segmentation model combining a hierarchical Transformer encoder with a lightweight decoder for efficient dense prediction. We adopt the B3 and B4 backbone variants, pretrained on ADE20K~\cite{zhou2019semantic}.

\textbf{Mask Classification Models.}
We consider two state-of-the-art semantic segmentation architectures for mask classification. 
\textit{MaskFormer}~\cite{cheng2021per}, inspired by DETR~\cite{carion2020end}, unifies semantic, instance, and panoptic segmentation under a mask-classification framework. We adopt base and large model variants with Swin Transformer~\cite{liu2021swin} backbones, with models pretrained on ADE20K and COCO-Stuff.
\textit{Mask2Former}~\cite{cheng2022masked} extends MaskFormer by introducing masked attention,  a variant of cross attention that iteratively attends within the region of the predicted mask for each query, in the transformer decoder, and leveraging multi-scale high-resolution features, improving efficiency and segmentation accuracy. 
We evaluate base and large versions with Swin Transformer backbones pretrained on ADE20K and COCO-Stuff.

\textbf{Implementation Details} 
\textit{Per-Pixel Models.}
All images are resized to $512\times512$  with standard ImageNet normalization. 
To stabilize hyperbolic computations, we clamp feature magnitudes and apply weight decay during training. Following \cite{desai2023hyperbolic}, we introduce learnable scaling parameters $\alpha_{txt}$ and $\alpha_{img}$ for text and image embeddings, initialized as the inverse of their mean feature norms.
For per-pixel models, a pretrained encoder (SegFormer or DeepLabV3) is coupled with a customized decoder head to handle class mismatch, feature alignment with text embeddings, and output interpolation. Sentence embeddings are generated using the BAAI/bge-m3 model; base and large variants yield similar performance. After lifting to the Lorentz model, image embeddings have shape $[B, H, W, d+1]$ and text embeddings $[C,d+1]$, producing logits $[B, H, W, C]$ via the negative Lorentzian inner product. Training uses AdamW with weight decay, employing a smaller learning rate for the encoder and a larger one for the decoder. The embedding dimension is set to $d=8$ for Cityscapes and Pascal VOC, and $d=64$ for ADE20K and COCO-Stuff. The entailment loss weight is $\lambda_w=0.5$.

\textit{Mask Classification Models.}
Mask2Former architectures by replacing the linear classifier with hyperbolic prototype matching in the Lorentz model to compute class query logits, and by replacing the Euclidean dot product with hyperbolic sample-to-sample similarity to compute mask query logits, as described in Sec .~\ref{sec:method}.

For class prototype construction, text prototypes are scaled by their mean norms to produce embeddings of unit tangent-space norm, which are then lifted via the exponential map. At unit tangent-space norm, the half-aperture angle evaluates to:
$
\mathrm{aper} \approx \sin^{-1}\!\left(\frac{2 \times 0.1}{\sinh(1)}\right) \approx 0.17 \text{ rad}.
$
After normalization and mapping, the mean geodesic distance between class prototypes is $4.1 \pm 0.2$, confirming well-separated class anchors in hyperbolic space. Noting that $\cosh^{-1}(2) \approx 1$, enforcing a unit-radius constraint on the anchor embeddings bounds the Lorentz inner product to the range [-2, -1].

For mask query logit computation, the shift and scale parameters of Eq.~\ref{eq:scale_shift} are set as follows. The angle shift is set to $b_a = 0.17$, consistent with the aperture value for a unit-norm Lorentzian tangent, such that embeddings near the cone boundary are mapped to zero before sigmoid activation. To ensure confident mask activations, the angle scale is set to $s_a = 0.02$, mapping the shifted exterior angle range of $(-\pi + 0.17, 0.17)$, which maps exterior angular similarities corresponding to $0 - 0.13$ radians to sigmoid outputs greater than $0.9$. The distance shift is set to $b_d = 1$, and the distance scale to $s_d = 0.1$, such that pixel embeddings with a shifted negative distance in the range $(0.22, 1)$, corresponding to a geodesic distance of at most $0.78$,  yield sigmoid activations above $0.9$. For Mask2Former, we extend the same hyperbolic similarity formulation to the masked attention module, enabling iterative mask refinement using hyperbolic operations. Training follows the standard MaskFormer optimization procedure, including Hungarian matching for query–ground-truth assignment, cross-entropy loss for classification, and focal and Dice losses for mask supervision.


\textbf{Evaluation Metrics:} 
We report the standard single-scale mean Intersection-over-Union (mIoU) for semantic segmentation, where IoU measures the spatial overlap between predicted and ground-truth masks, averaged over all classes without test-time augmentation. To assess hierarchical consistency, we additionally introduce sibling and cousin mIoU variants based on WordNet hypernym–hyponym relations (e.g., \textit{car}–\textit{bus} under \textit{vehicle})~\cite{atigh2022hyperbolic}.
All per-pixel experiments use single-stage inference on $512 \times 512$ images with SegFormer and DeepLabV3 backbones across ADE20K, COCO-Stuff, Pascal-VOC, and Cityscapes. For mask-classification architectures (MaskFormer and Mask2Former), experiments are conducted on $640 \times 640$ images using ADE20K and COCO-Stuff pretrained models, evaluating both base and large variants.
Beyond standard mIoU, we assess cross-dataset generalization, hierarchical consistency, uncertainty estimation, and boundary delineation, and provide qualitative visualizations with comparisons to relevant baselines. We omit SegFormer fine-tuning on Pascal-VOC due to its limited size, and do not fine-tune MaskFormer models on Cityscapes or Pascal-VOC for the same reason.

\subsection{Semantic Segmentation Performance}

\begin{figure}
    \centering
    \includegraphics[width=1\linewidth]{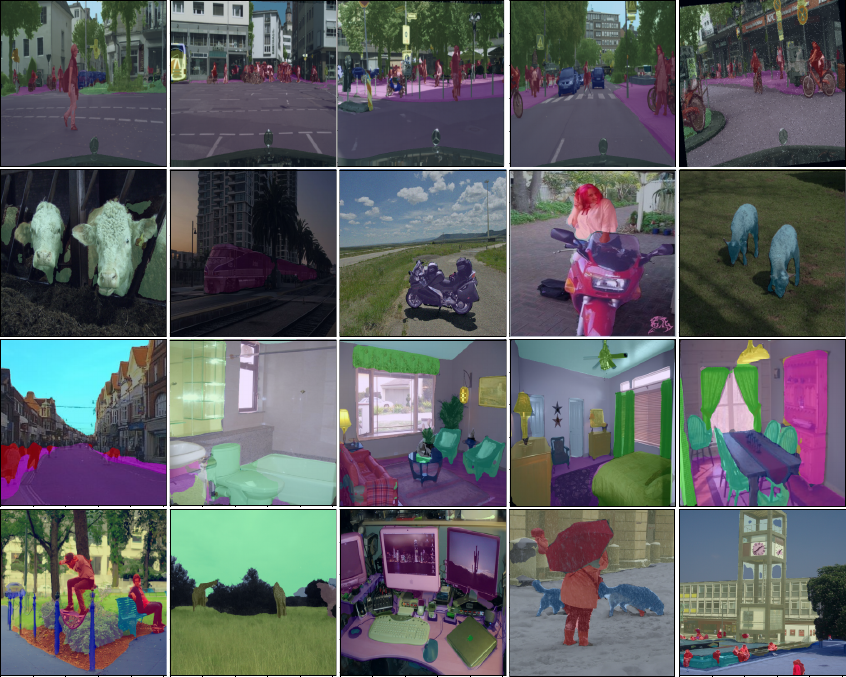}
    \caption{Qualitative segmentation results across Citiscapes, Pascal-VOC, Ade20k, and COCO-Stuff test samples. }
    \label{fig:all_map}
\end{figure}

\begin{figure}
    \centering
    \includegraphics[width=1\linewidth]{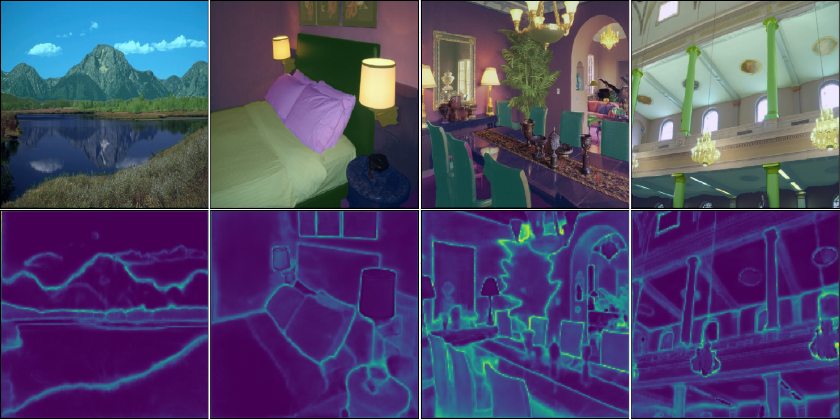}
    \caption{Qualitative segmentation and angle-based boundary detection by mask2former on ADE20K data samples. }
    \label{fig:all_map_mf}
\end{figure}

\begin{figure}
    \centering
    \includegraphics[width=1\linewidth]{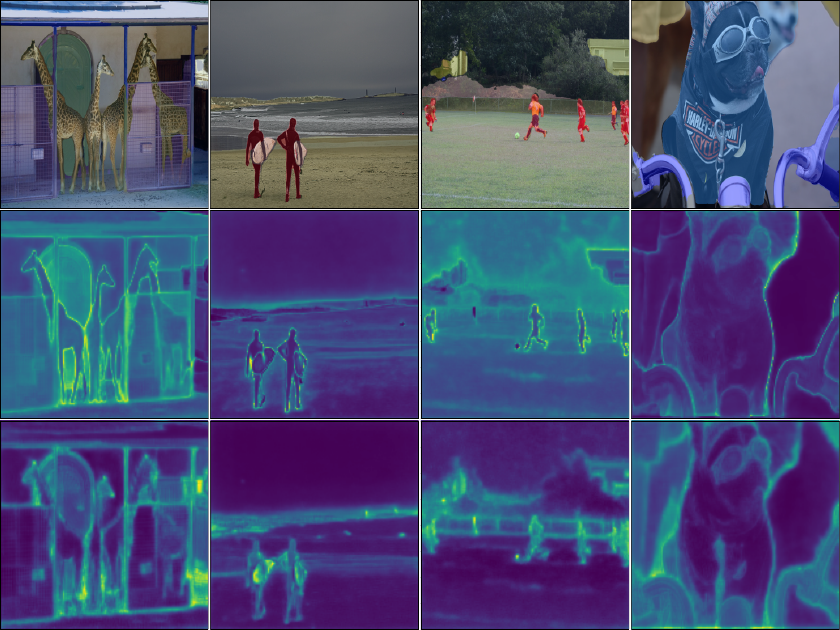}
    \caption{Qualitative segmentation and angle-based boundary detection by mask2former on Coco-stuff data samples. 1st row semantic mask, 2nd row angle based uq, 3rd row, distance based uq.}
    \label{fig:all_map_mf_coco}
\end{figure}

Our Lorentz semantic segmentation framework supports two inference strategies. The first follows the conventional approach of assigning each pixel to the class prototype with the minimal geodesic distance, i.e., $\underset{i}{\operatorname{argmin}} \; d_{\mathcal{L}}(\bm x_i,\bm y_j)$ for the $j$-th pixel embedding. Alternatively, we introduce a hierarchical inference strategy that selects the prototype with the minimal exterior angle, $\underset{i}{\operatorname{argmin}} \; \text{ext}(\bm x_i,\bm y_j)$ (Eq.~\ref{eq:ext_angle}), enforcing an \texttt{is-a} relationship consistent with entailment-based embeddings. Both approaches yield highly similar segmentation maps; unless otherwise stated, we report results using the distance-based method.

For mask-classification models, query class logits and mask query logits are computed using the proposed hyperbolic similarity formulation. During inference, the predicted class logits $(N, C)$ and mask logits $(h, w, N)$ are combined using the standard MaskFormer strategy to obtain the final segmentation map $(h, w, C)$.

We evaluate our approach using PCA-reduced sentence embeddings with dimension $d=64$ for ADE20K and COCO-Stuff, and $d=8$ for Pascal-VOC and Cityscapes. Our text-supervised Lorentzian embeddings achieve accuracy comparable to Euclidean embedding models while remaining competitive with existing text-supervised semantic segmentation methods, despite using substantially lower-dimensional prototypes (Table~\ref{tab:full_ss_miou}). Representative qualitative results across datasets and models are shown in the main paper, with additional experiments and ablations provided in the supplementary material.

\begin{table}[]
    \centering
    \resizebox{\columnwidth}{!}{
    \begin{tabular}{lcccc}
    Method                 & Pascal-VOC~\cite{everingham2010pascal} & Cityscapes~\cite{cordts2016cityscapes} & ADE20K~\cite{zhou2019semantic} & COCO-Stuff~\cite{caesar2018coco}\\ \hline
    DeepLabV3               & 85.7  & 80.1        & 42.7     & 38       \\
    SegFormer (B3)          &     & 81.7        & 49.4     & 45.5       \\
    SegFormer (B4)          &     & 82.3        & 50.3     & 46.5       \\
    MaskFormer Swin B       &     &             &  51.1       &   51.1  \\
    MaskFormer Swin L       &     &             &  54.1       &   52.7  \\
    Mask2Former Swin L       &     &             &  56.1       &   \\
    MTA-clip (R101)         &     &           & 49.8     &          \\
    MTA-clip (ViT-B)        &     & 82.1        & 52.3     &          \\
    DenseClip (R101)        &     & 81        & 45.1       &          \\
    DenseClip (ViT-B)       &     &           & 50.6     &          \\
    MTA-clip (MiT-B4)       &     &           & 51.7     &          \\ \hline
    Ours (Deeplab-R101)         & 84.8  & 80.6     & 43.4    & 41.5     \\
    Ours (SegFormer-B3)      &     & 83.7      & 50.1   & 47.2     \\ 
    Ours (MaskFormer SB)       &     &             &  51.8       &   48.6  \\
    Ours (MaskFormer SL)       &     &             &  54.5       &   50.1  \\
    Ours (Mask2Former SL)       &     &           &  56.7       &   51.6 \\\hline
    \end{tabular}}
    \caption{Quantitative results on ADE20K, COCO-Stuff, Pascal-VOC, and Cityscapes. We compare mIoU for Lorentzian embeddings with relevant baselines. Our framework achieves competitive or superior performance across datasets with minimal computational overhead.}
    \label{tab:full_ss_miou}
\end{table}

\subsection{Uncertainty and confidence}

\begin{figure}
    \centering
    \includegraphics[width=1\linewidth]{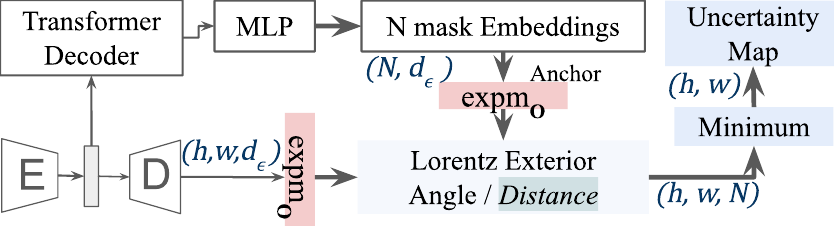}
    \caption{Single Pass Uncertainty Estimation using Mask classification models.}
    \label{fig:mf_uq}
\end{figure}

\begin{figure}
    \centering
    \includegraphics[width=1\linewidth]{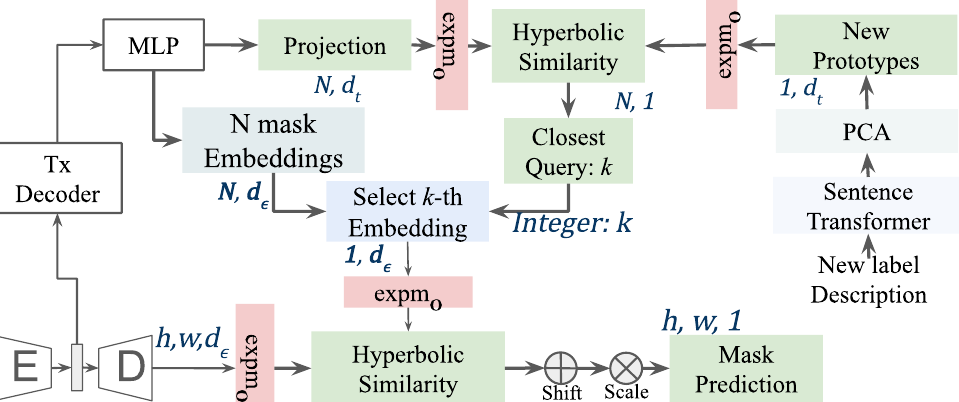}
    \caption{Text query-based segmentation using mask classification models.}
    \label{fig:mf_ti}
\end{figure}

Uncertainty Estimation for the Maskformer, Zero-shot, New Architecture diagram.
Results: new table diagram.

Hyperbolic embeddings offer geometrically interpretable predictions, making them particularly attractive for uncertainty quantification~\cite{nickel2017poincare, nickel2018learning, chen2023hyperbolic}. Our framework enables pixel-level uncertainty estimation without requiring ground-truth labels, providing insight into both prediction confidence and semantic boundary structure. We identify two complementary families of uncertainty measures in Lorentz space: \textit{length-based} and \textit{angle-based}.

\textbf{Connection with Distance-based Uncertainty.}
Following~\cite{atigh2022hyperbolic}, we investigate whether Lorentz embeddings preserve the length-order properties of Poincaré embeddings under the isometric mapping. The Lorentz-to-Poincaré projection (Eq.~\ref{eq:l2p}) yields uncertainty maps based on radial distance from the origin, where lower radius corresponds to higher uncertainty, with elevated uncertainty consistently concentrated along semantic boundaries — consistent with prior findings~\cite{khrulkov2020hyperbolic, atigh2022hyperbolic}. Furthermore, both the Lorentz-to-Poincaré and Euclidean-to-Lorentz mappings (Eq.~\ref{eq:l_map_O}) preserve length order, permitting direct use of the Euclidean norm $\|\bm{x}'\|$ or the time component $x_0$ as a \textit{prototype-free} uncertainty metric, consistent with the findings of~\cite{nickel2018learning}.

\textbf{Novel Angle-based Uncertainty.}
We introduce a complementary \textit{angle-based} uncertainty measure enabled by the entailment loss, which constrains pixel embeddings to lie within the entailment cone of their label prototypes or mask embeddings. 
For per-pixel classification models, we compute the minimal exterior angle between each pixel embedding and the set of class prototypes, $\underset{i}{\min}\,\text{ext}(\bm x_i,\bm y_j)$. Larger angles indicate stronger deviations from the prototype entailment cone and therefore lower confidence in the predicted class. This \textit{prototype-aware} metric naturally reflects the hierarchical \texttt{is-a} structure encoded by the entailment formulation \cite{vendrov2015order, nickel2018learning}.
For mask-classification models, we compute the exterior angle between each pixel embedding and the set of $N$ mask embeddings predicted by the decoder. For each of the $(h,w)$ pixels, we evaluate $N$ angles relative to the mask embeddings and take the minimum as the uncertainty value (Fig.~\ref{fig:mf_uq}). This value reflects how strongly the pixel aligns with its closest mask anchor (Fig.~\ref{fig:all_map_mf}). Pixels exhibiting higher minimum exterior angles tend to align with multiple mask queries, indicating boundary ambiguity and elevated prediction uncertainty.
While distance-based similarity can also be used for this purpose, we empirically observe that the angle-based measure provides clearer boundary localization in mask classification models.

Empirically, uncertain pixels consistently fall outside the expected entailment cone, violating order-embedding constraints and admitting a geometric interpretation of epistemic uncertainty in hyperbolic space. The angle-based measure complements distance-based uncertainty by offering a prototype-conditioned perspective on prediction confidence.

\textbf{Finding Object Confidence.}
We localize confident regions by computing classwise confidence using averaged features in the Klein model for efficiency. Pixel embeddings of a selected class are mapped to Klein space \ref{eq:l2k}, averaged via the Einstein midpoint \ref{eq:einsten_mp}, and mapped back to Lorentz space to obtain the class mean $\bm y_m$. Confidence heatmaps are then generated using either $\exp(-d_\mathcal{L}(\bm y_j, \bm y_m))$ or exterior angle deviations from $\bm y_m$, capturing both feature-level similarity and geometric alignment with the class prototype or mask embeddings cones.

We hypothesize that distance-based measures primarily reflect \textit{aleatory uncertainty}, capturing variability in feature embeddings and inherent data noise, whereas angle-based measures capture \textit{epistemic uncertainty}, reflecting the confidence of embeddings relative to the semantic prototypes and hierarchy. A formal theoretical characterization of this distinction is beyond the scope of this work and is left for future study.

\subsection{Beyond trained-set performance:}

\begin{figure}
    \centering
    \includegraphics[width=1\linewidth]{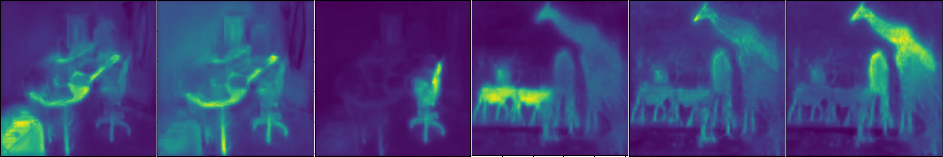}
    \caption{Hierarchical Class retrieval example from COCO-Stuff. From Left, retrieved map by table, furniture, and chair classes for the same images. Similarly, retrieved the map by Zebra, Novel text: \textit{Wild Animal}, and Giraffe classes. }
    \label{fig:coco_hier}
\end{figure}

\begin{figure}
    \centering
    \includegraphics[width=1\linewidth]{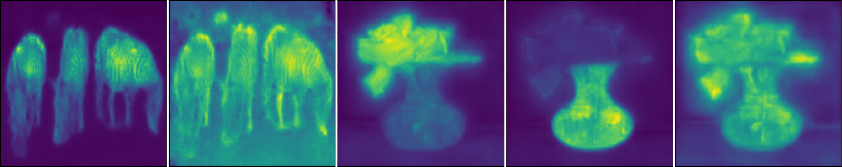}
    \caption{Retrieving Class map using relevant classes. From left, Zebra image using Zebra and Horse query. An image with the query of flower, vase, and potted flower.}
    \label{fig:coco_cross_retrieval}
\end{figure}

\begin{figure}
    \centering
    \includegraphics[width=1\linewidth]{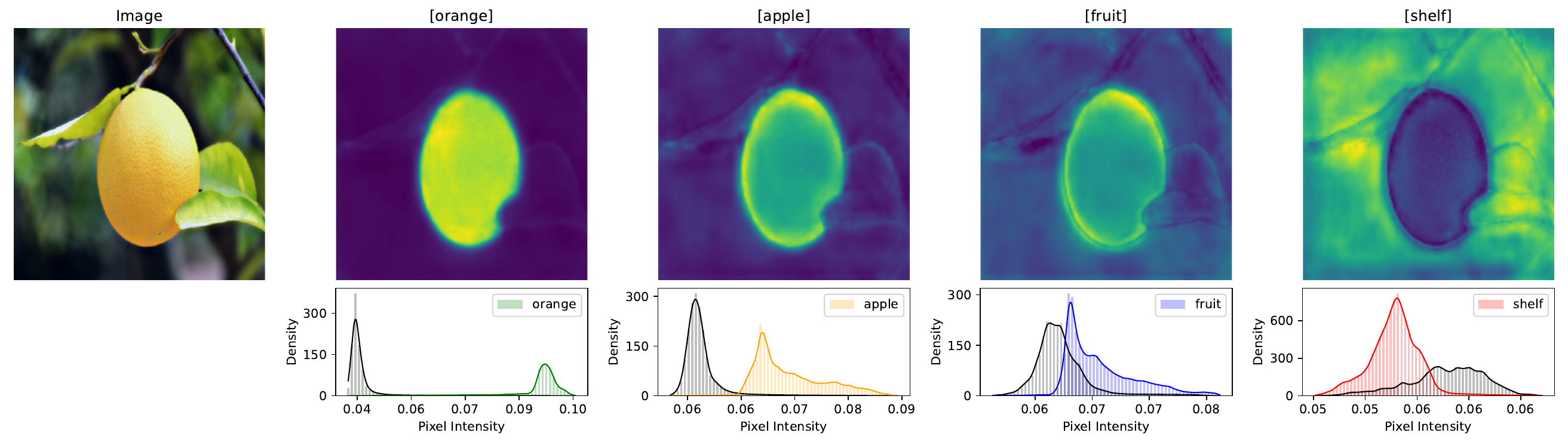}
    \caption{Image mask captured by different keywords. We point to the corresponding histogram; the accurate class text has the highest confidence, and so on (more results in the supplementary section).  }
    \label{fig:placeholder}
\end{figure}
We investigate the ability of our Lorentz embedding to generalize toward hierarchical semantic retrieval and zero-shot learning beyond the trained label set. For unseen classes, we follow the same label processing pipeline: we provide semantic descriptions, compute sentence embeddings, and apply the PCA transformation trained on labeled classes to obtain low-dimensional prototypes for novel classes.

\textbf{Hierarchical Retrieval.}
We analyze the \textit{hierarchical} behavior of our model for hypernym–hyponym relationships and visually similar classes (e.g., car–truck, goat–sheep). Leveraging the entailment-based training objective, our Lorentz embedding naturally captures hierarchical structures, enabling the retrieval of semantically related classes from text queries. For example, “vehicle” retrieves bus and car, “flying things” retrieves aeroplane, frisbee, kite, and bird, while cow, sheep, and goat cluster under “animal.” WordNet hypernym queries further confirm alignment with linguistic hierarchies. We provide sample results for COCO-Stuff in figure \ref{fig:coco_hier}. 

\textbf{Text-based Semantic Retrieval.}
For per-pixel classification models, an input text description is embedded and projected into Lorentz space, and pixels are retrieved by nearest-prototype matching, as illustrated in Fig.~\ref{fig:overview}(c). 
For mask-classification models (Fig.~\ref{fig:mf_ti}(c)), the input text embedding is matched against the $N$ class query embeddings in hyperbolic space via the sample-to-prototype similarity, identifying the closest query. The corresponding mask embedding is then used to generate the retrieved semantic mask for the text query via Lorentz similarity.
For instance, “two-wheeled vehicles” retrieves bicycle and motorcycle, “wild animals” retrieves zebra, giraffe, and bear, “domestic animals” retrieves dog, horse, and cat, and “indoor furniture” retrieves cabinet, desk, and chair (Fig. \ref{fig:coco_hier}). Top-$k$ retrieval consistently yields semantically coherent results, such as horse–zebra, plane–bird, and cow–sheep–goat clusters.

\textbf{Cross-label Retrieval.}
Our model can retrieve semantically related classes using similar text queries, demonstrating its ability to capture structural relevance (Fig. \ref{fig:coco_cross_retrieval}). For example, a bird mask may be retrieved by an “aeroplane” query, a zebra mask by a “horse” query, or a sheep mask by a “goat” query. However,  such cases exhibit a large exterior angle $\text{ext}(\bm{x}_{bird}, \bm{y}_{\text{plane}})$, indicating semantic deviation and high epistemic uncertainty. Both distance- and angle-based metrics capture this behavior, highlighting the robustness of our uncertainty estimation.


\textbf{Zero-shot Performance.}
\begin{figure}
    \centering
    \includegraphics[width=1\linewidth]{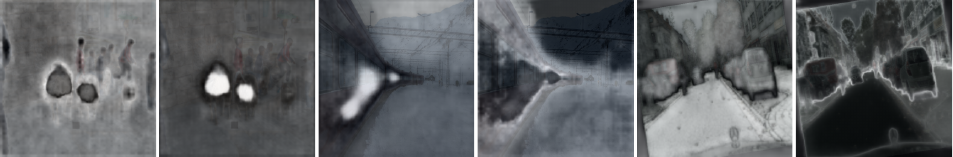}
    \caption{Higher Uncertainty for the zero-shot retrieved objects (distance \& angle). Sheep, train, and Rider pixels have higher uncertainty.}
    \label{fig:zs_uncertainty}
\end{figure}
We investigate the zero-shot segmentation performance of our model using visual descriptions and the Lorentzian \texttt{is-a} hierarchical relationship. Specifically, we evaluate scenarios where certain classes are withheld during training under two settings: (i) the Pascal-VOC 15/5 split following \cite{atigh2022hyperbolic, xian2019semantic}, and (ii) Cityscapes zero-shot, where a COCO-Stuff–pretrained DeepLabV3 model omits bicycle, sidewalk, rider, and fence classes. In inference, textual descriptions of unseen classes are provided to infer semantic masks. Results (Tab. \ref{tab:zs_table} and Fig.~\ref{fig:zs_uncertainty}) demonstrate that our model achieves competitive zero-shot segmentation performance, successfully retrieving novel classes while naturally assigning higher uncertainty to regions containing unseen semantics.

These experiments demonstrate that text supervision in Lorentz embeddings enables hierarchical embedding learning, semantic generalization, and zero-shot recognition.

\begin{table}[h!]
    \centering
    \resizebox{\columnwidth}{!}{
    \begin{tabular}{l|cccc}
    \hline
         & CE ($\mathbb{R}^n$) & $\mathbb{R}^n$ (proto) & Poincaré\cite{atigh2022hyperbolic} & Ours ($\mathbb{R}^{1,n}$)  \\     \hline 
    mIOU &   2.5  &    18.6       &    34.9   &    36.4 (8-d) \\ 
    Uncertainty &  \ding{55} & \ding{55} & \ding{51} & \ding{51} \\ 
'\texttt{is-a}' relation & \ding{55} & \ding{55} & \ding{55} & \ding{51} \\
    Computation &  &  & High &  \\ 
    Text-query & \ding{55} & \ding{51} & \ding{55} & \ding{51} \\ 
    Hierarchical & \ding{55} & \ding{55} & \ding{51} & \ding{51} \\ \hline
    \end{tabular}
    }
    \caption{Zero-shot result comparison for Pascal-VOC using DeepLabV3 (resnet-101) backbone. 
    R stands for Euclidean network. Comparison with other benefits.}
    \label{tab:zs_table}
\end{table}

\subsection{Loss Landscape Geometry}

\begin{figure}
    \centering
    \includegraphics[width=1\linewidth]{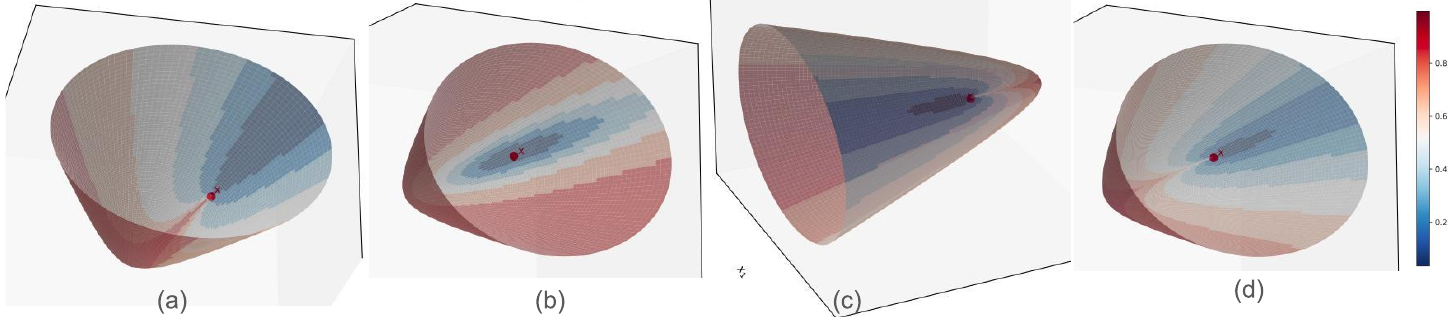}
    \caption{Visualization of Loss Geometry for a $\mathbb{R}^{1,2}$ Lorentz Model (a) entailment loss, (b) Geodesic distance loss, (c) and (d) Weighed sum of the losses from two sides}
    \label{fig:loss_geom_2d}
\end{figure}
We empirically analyze the loss geometry of our Lorentz framework to understand its learning dynamics. Fig.~\ref{fig:loss_geom_2d} visually depicts the geodesic distance loss and the entailment loss in a $\mathbb{R}^{1,2}$ Lorentz model.
The geodesic distance loss (Fig. \ref{fig:loss_geom_2d}b) acts as a hyperbolic "geodesic circles" around the prototype, analogous to a Euclidean circle. It enforces proximity to class prototypes without directional constraints—resulting in a geodesic circle loss surface. 
The entailment cone loss (Fig. \ref{fig:loss_geom_2d}a) provides a directional inductive bias by encouraging embeddings to fall within the entailment cone defined by prototypes' location, consistent with the hierarchical \texttt{is-a} relationship. Their weighted sum creates a complex loss landscape (\ref{fig:loss_geom_2d}(c,d)), which encourages the embedding to fall in the entailment cone while being near the prototypes. 

\begin{figure}
    \centering
    \includegraphics[width=1\linewidth]{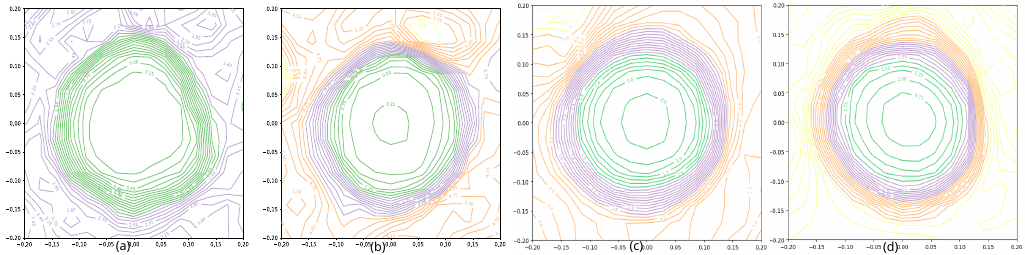}
    \caption{Loss surface Visualization for DeepLabV3 cityscape (a) $\mathcal{H}$ (b) R, SegFormer COCO-Stuff (c) $\mathcal{H}$ (d) R. Empirical visualization shows that Lorentz model consistently achieves smoother loss than Euclidean.  }
    \label{fig:loss_vis}
\end{figure}

Following \textit{filter normalization} \cite{li2018visualizing}, we perturb model weights along two random directions to visualize the loss surfaces. Our empirical observation reveals that the Lorentz model converges to flatter minima than its Euclidean counterpart (fig \ref{fig:loss_vis}), aligning with the well-known link between flat minima and improved generalization \cite{keskar2016large, dinh2017sharp}, and explaining its strong empirical performance. A detailed analysis of the loss gradients lies beyond the scope of this paper and will be explored in future work. In particular, we plan to investigate why hyperbolic embeddings exhibit such behavior and how the choice of Lorentz hyperbolic space and loss functions influences the optimization dynamics.

\subsection{Ablation Studies}
We conduct extensive studies on different components.

\textbf{Text Encoding.}
We evaluate multiple sentence encoding strategies, comparing distributional embeddings (WordNet + GloVe) with contextual embeddings obtained from transformer encoders. Contextual embeddings consistently outperform distributional counterparts, as they better capture semantic information. We further examine the effect of varying text descriptions and observe that adding or modifying words has minimal impact on the resulting embeddings since sentence encoders produce semantically consistent representations.

\textbf{Embedding Dimensionality.}
We analyze the impact of embedding dimensionality on performance. While Lorentz embeddings maintain stable performance across dimensions, Euclidean embeddings degrade in segmentation performance as the dimensionality decreases. Notably, the Lorentz model achieves competitive performance on both segmentation architectures even with lower dimensions, showcasing its geometric efficiency.

\textbf{Model Architectures.}
We test scalability across architectures by using DeepLabV3 (ResNet-50, ResNet-101), SegFormer (MiT-B3, MiT-B4), Maskformers (base and Large models). Our model consistently performs well across all datasets and architectures, demonstrating robustness and general applicability. We observe similar results for mask classification models. 

\textbf{Loss Component Analysis.}
We isolate the contribution of each loss term. Without the entailment loss, embeddings converge toward prototypes from arbitrary directions, losing the semantic \texttt{is-a} hierarchy encoded via order embeddings and eliminating angle-based uncertainty estimation. Nevertheless, hyperbolic distance-only training still matches Euclidean ones under the same conditions.

\textbf{Mask former Modification Analysis.}
We investigate the contribution of maskformer angle and distance components. The angle entailment and hyperbolic Lorentz distances combine to capture the \texttt{is-a} hierarchy. While generating the mask, both components provide the relation with the pixels embedding with the text prototypes in a sample-to-sample fashion instead of directly in a sample-to-prototype fashion in pixel classification models.

\textbf{Euclidean Baseline.}
Replacing hyperbolic space with Euclidean ($\mathbb{R}^n$) degrades performance for zero-shot retrieval, where Euclidean prototypes ($\mathbb{R}^n$ (Proto)) fail to preserve hierarchical relationships, dropping zero-shot performance (Tab. \ref{tab:zs_table}). Further, text-supervised (R(proto)) or traditional cross-entropy supervised ($\mathbb{R}^n$)  models do not provide uncertainty estimates. In contrast, our Lorentz framework on different segmentation architectures consistently captures semantic hierarchies and uncertainty via its geometric inductive bias, confirming the necessity of hyperbolic geometry for robust generalization.




\section{Conclusion}
We presented a text-supervised hyperbolic semantic segmentation framework in the Lorentz model that enables scalable pixel-level learning across both per-pixel and mask-classification architectures. By introducing an entailment cone loss, our approach models each pixel or mask embedding as an instance of its corresponding class prototype, capturing hierarchical \texttt{is-a} relationships while remaining fully compatible with modern segmentation pipelines. Beyond strong empirical performance across multiple benchmarks, our analysis shows that hyperbolic representations provide structured inductive biases that improve hierarchical consistency, zero-shot generalization, uncertainty estimation, and representational efficiency, particularly in low-dimensional regimes. Furthermore, our findings suggest that the benefits of hyperbolic geometry extend beyond representation capacity to influence optimization dynamics and learning behavior. These results point to several promising directions for future work, including integrating hyperbolic geometry with large-scale multimodal pretraining, developing a deeper theoretical understanding of optimization in curved spaces, and extending these ideas to broader structured prediction tasks.

\textbf{Acknowledgement:} This work has been partially supported by ONR Grant \#N00014-23-1-2119, U.S. Army Grant \#W911NF2120076,NSF CAREER Grant \#1750936, NSF CNS EAGER Grant \#2233879, NSF IIS Grant \#2509680, and U.S. Army Grant \#W911NF2410367.

{\small
\bibliographystyle{ieee_fullname}
\bibliography{egbib}
}

\onecolumn
\section*{Derivative Computations for Various Functions}

\subsection*{Dot Product}

Given:
\begin{itemize}
    \item $\mathbf{x} = [x_1, x_2, \ldots, x_d]$
    \item $\mathbf{y}_i = [y_{i1}, y_{i2}, \ldots, y_{id}]$
    \item $f_i = \mathbf{x} \cdot \mathbf{y}_i = \sum_{k=1}^{d} x_k y_{ik}$
\end{itemize}

Derivative:
\[
\frac{\partial f_i}{\partial x_j} = \frac{\partial}{\partial x_j} \left( \sum_{k=1}^{d} x_k y_{ik} \right) = y_{ij}
\]

\subsection*{Euclidean Distance}

Given:
\begin{itemize}
    \item $\mathbf{x} = [x_1, x_2, \ldots, x_d]$
    \item $\mathbf{y}_i = [y_{i1}, y_{i2}, \ldots, y_{id}]$
    \item $f_i = \|\mathbf{x} - \mathbf{y}_i\|_2 = \sqrt{\sum_{j=1}^{d} (x_j - y_{ij})^2}$
\end{itemize}

Let $g_i = \|\mathbf{x} - \mathbf{y}_i\|_2^2 = \sum_{k=1}^{d} (x_k - y_{ik})^2$, so $f_i = \sqrt{g_i}$.

Using chain rule:
\[
\frac{\partial f_i}{\partial x_j} = \frac{1}{2} g_i^{-1/2} \cdot \frac{\partial g_i}{\partial x_j}
\]

Compute $\frac{\partial g_i}{\partial x_j}$:
\[
\frac{\partial g_i}{\partial x_j} = 2(x_j - y_{ij})
\]

Thus:
\[
\frac{\partial f_i}{\partial x_j} = \frac{1}{2} \cdot \frac{1}{\sqrt{g_i}} \cdot 2(x_j - y_{ij}) = \frac{x_j - y_{ij}}{\|\mathbf{x} - \mathbf{y}_i\|_2}
\]

\subsection*{Cosine Similarity}

Given:
\begin{itemize}
    \item $\mathbf{x} = [x_1, x_2, \ldots, x_d]$
    \item $\mathbf{y}_i = [y_{i1}, y_{i2}, \ldots, y_{id}]$
    \item $f_i = \frac{\mathbf{x} \cdot \mathbf{y}_i}{\|\mathbf{x}\| \|\mathbf{y}_i\|}$
\end{itemize}

Let $u = \mathbf{x} \cdot \mathbf{y}_i$ and $v = \|\mathbf{x}\|$:
\[
\frac{\partial f_i}{\partial x_j} = \frac{1}{\|\mathbf{y}_i\|} \cdot \frac{\partial}{\partial x_j} \left( \frac{u}{v} \right)
\]

Using quotient rule:
\[
\frac{\partial}{\partial x_j} \left( \frac{u}{v} \right) = \frac{ \frac{\partial u}{\partial x_j} \cdot v - u \cdot \frac{\partial v}{\partial x_j} }{v^2}
\]

Compute partials:
\begin{align*}
\frac{\partial u}{\partial x_j} &= y_{ij} \\
\frac{\partial v}{\partial x_j} &= \frac{x_j}{\|\mathbf{x}\|} = \frac{x_j}{v}
\end{align*}

Substitute:
\[
\frac{\partial}{\partial x_j} \left( \frac{u}{v} \right) = \frac{ y_{ij} v - u \cdot \frac{x_j}{v} }{v^2} = \frac{ y_{ij} v^2 - u x_j }{v^3}
\]

Final result:
\[
\frac{\partial f_i}{\partial x_j} = \frac{ y_{ij} \|\mathbf{x}\|^2 - (\mathbf{x} \cdot \mathbf{y}_i) x_j }{ \|\mathbf{y}_i\| \|\mathbf{x}\|^3 }
\]

\subsection*{Normalized Vectors Case}

If $\|\mathbf{x}\| = 1$ and $\|\mathbf{y}_i\| = 1$, then:
\[
f_i = \mathbf{x} \cdot \mathbf{y}_i = \sum_{k=1}^{d} x_k y_{ik}
\]
\[
\frac{\partial f_i}{\partial x_j} = y_{ij}
\]

\section*{Hyperbolic operations}

\subsection*{Hyperbolic Cosine with Lorentz Product}

Given:
\begin{itemize}
    \item $\mathbf{x} = [x_1, x_2, \ldots, x_d]$
    \item $\mathbf{y}_i = [y_{i1}, y_{i2}, \ldots, y_{id}]$
    \item Lorentz product: $\langle \mathbf{x}, \mathbf{y}_i \rangle_L = -x_0 y_{i0} + \sum_{j=1}^{d} x_j y_{ij}$
    \item $x_0 = \sqrt{1 + \|\mathbf{x}\|^2}$, $y_{i0} = \sqrt{1 + \|\mathbf{y}_i\|^2}$
    \item $f_i^d = \cosh^{-1}(-\langle \mathbf{x}, \mathbf{y}_i \rangle_L)$
\end{itemize}

Let $u = -\langle \mathbf{x}, \mathbf{y}_i \rangle_L = x_0 y_{i0} - \sum_{j=1}^{d} x_j y_{ij}$

Then:
\[
f_i = \cosh^{-1}(u)
\]

Using the chain rule:
\[
\frac{\partial f_i^d}{\partial x_j} = \frac{1}{\sqrt{u^2 - 1}} \cdot \frac{\partial u}{\partial x_j}
\]

Compute $\frac{\partial u}{\partial x_j}$:
\[
\frac{\partial u}{\partial x_j} = y_{i0} \cdot \frac{\partial x_0}{\partial x_j} - y_{ij}
\]

Since $\frac{\partial x_0}{\partial x_j} = \frac{x_j}{x_0}$:
\[
\frac{\partial u}{\partial x_j} = y_{i0} \cdot \frac{x_j}{x_0} - y_{ij}
\]

Final result:
\[
\frac{\partial f_i^d}{\partial x_j} = \frac{-( y_{ij} - \dfrac{y_{i0}}{x_0} x_j )}{ \sqrt{ \left( x_0 y_{i0} - \sum\limits_{k=1}^{d} x_k y_{ik} \right)^2 - 1 }}
\]

\subsection*{Arccosine with Lorentz Product}

Given:
\begin{itemize}
    \item $f_i^a = \arccos\left( \frac{x_0 + y_{i0} \langle \mathbf{x}, \mathbf{y}_i \rangle_L}{\|\mathbf{y}_i\| \sqrt{\langle \mathbf{x}, \mathbf{y}_i \rangle_L^2 - 1}} \right)$
\end{itemize}

Let:
\begin{align*}
L &= \langle \mathbf{x}, \mathbf{y}_i \rangle_L \\
N &= x_0 + y_{i0} L \\
D &= \sqrt{L^2 - 1} \\
A &= \frac{N}{\|\mathbf{y}_i\| D}
\end{align*}

Then:
\[
f_i^a = \arccos(A)
\]

Using the chain rule:
\[
\frac{\partial f_i^a}{\partial x_j} = -\frac{1}{\sqrt{1 - A^2}} \cdot \frac{\partial A}{\partial x_j}
\]

Compute $\frac{\partial A}{\partial x_j}$:
\[
\frac{\partial A}{\partial x_j} = \frac{1}{\|\mathbf{y}_i\|} \cdot \frac{\partial}{\partial x_j} \left( \frac{N}{D} \right)
\]

Using the quotient rule:
\[
\frac{\partial}{\partial x_j} \left( \frac{N}{D} \right) = \frac{D \cdot \frac{\partial N}{\partial x_j} - N \cdot \frac{\partial D}{\partial x_j}}{D^2}
\]

Since $D^2 = L^2 - 1$ and $\frac{\partial D}{\partial x_j} = \frac{L}{D} \frac{\partial L}{\partial x_j}$:
\[
\frac{\partial A}{\partial x_j} = \frac{1}{\|\mathbf{y}_i\| \sqrt{L^2 - 1}} \left( \frac{\partial N}{\partial x_j} - \frac{N L}{L^2 - 1} \frac{\partial L}{\partial x_j} \right)
\]

Compute partials:
\begin{align*}
\frac{\partial L}{\partial x_j} &= y_{ij} - \frac{y_{i0} x_j}{x_0} \\
\frac{\partial N}{\partial x_j} &= \frac{x_j}{x_0} + y_{i0} \left( y_{ij} - \frac{y_{i0} x_j}{x_0} \right) = y_{i0} y_{ij} - \frac{\|\mathbf{y}_i\|^2 x_j}{x_0}
\end{align*}

Substitute (formulation 1):
\[
\frac{\partial A}{\partial x_j} = \frac{1}{\|\mathbf{y}_i\| \sqrt{L^2 - 1}} \left[ y_{ij} \left( y_{i0} - \frac{N L}{L^2 - 1} \right) + \frac{x_j}{x_0} \left( -\|\mathbf{y}_i\|^2 + \frac{N L y_{i0}}{L^2 - 1} \right) \right]
\]

Final result:
\[
\frac{\partial f_i^a}{\partial x_j} = -\frac{1}{\sqrt{1-A^2}} \cdot \frac{1}{\|\mathbf{y}_i\| \sqrt{L^2-1}} \left[ y_{ij} \left( y_{i0} - \frac{N L}{L^2-1} \right) + \frac{x_j}{x_0} \left( - \|\mathbf{y}_i\|^2 + \frac{N L y_{i0}}{L^2-1} \right) \right]
\]

Substitute (formulation 2):
\begin{align*}
\frac{\partial A}{\partial x_j} = \frac{1}{\|\mathbf{y}_i\| \sqrt{L^2 - 1}} \left[\frac{x_j}{x_0} + y_{i0} \left( y_{ij} - \frac{y_{i0} x_j}{x_0} \right) - \frac{N L}{L^2 - 1} \left(y_{ij} - \frac{y_{i0} x_j}{x_0} \right) \right] \\
= \frac{1}{\|\mathbf{y}_i\| \sqrt{L^2 - 1}} \left[\frac{x_j}{x_0} + \left(y_{i0} - \frac{N L}{L^2 - 1}  \right) \left(y_{ij} - \frac{y_{i0} x_j}{x_0} \right) \right] \\
= \frac{1}{\|\mathbf{y}_i\| \sqrt{L^2 - 1}} \left[\frac{x_j}{x_0} + \left(\frac{y_{i0}(L^2-1) - (x_0 + y_{i0} L ) L}{L^2 - 1}  \right) \left(y_{ij} - \frac{y_{i0} x_j}{x_0} \right) \right] \\
= \frac{1}{\|\mathbf{y}_i\| \sqrt{L^2 - 1}} \left[\frac{x_j}{x_0} - \left(\frac{ y_{i0} + x_0L}{L^2 - 1}  \right) \left(y_{ij} - \frac{y_{i0} x_j}{x_0} \right) \right] 
\end{align*}

Final result:
\[
\frac{\partial f_i^a}{\partial x_j} = \frac{1}{\sqrt{1-A^2}} \cdot \frac{1}{\|\mathbf{y}_i\| \sqrt{L^2 - 1}} \left[- \frac{x_j}{x_0} + \left(\frac{ y_{i0} + x_0L}{L^2 - 1}  \right) \left(y_{ij} - \frac{y_{i0} x_j}{x_0} \right) \right]
\]

\subsection*{Angle Between $\nabla f^d$ and $\nabla f^a$}

Here,$\nabla f^* = [ \frac{\partial f*}{\partial x_j}]$. We simplify notation by replacing the $y_i$ and $f_i$ with $y$ and $f_i$. The dot product ($\nabla f^d \cdot (\nabla f^a)^T$) sign depends only on the sign of the following term, as the multiplicative terms are positive. 

\begin{align*}
\nabla f^d \cdot (\nabla f^a)^T = & \sum_j \left [-(y_{j} - \dfrac{y_{0}}{x_0} x_j)(- \frac{x_j}{x_0} + \left(\frac{ y_{0} + x_0L}{L^2 - 1}  \right) \left(y_{j} - \frac{y_{0} x_j}{x_0} \right)) \right ]\\
& = -\sum_j \left [  \left(\frac{ y_{0} + x_0L}{L^2 - 1}  \right) \left(y_{j} - \frac{y_{0} x_j}{x_0} \right)^2 - (y_{j} - \dfrac{y_{0}}{x_0} x_j)(\frac{x_j}{x_0}) \right ]    \\
& = -\left(\frac{ y_{0} + x_0L}{(L^2 - 1)x_0^2}  \right) \left ( \sum_j  \left(y_{j}x_0 -y_{0} x_j \right)^2 - (L^2-1) \right ); \text{derived later} 
\end{align*}

Close inspection of the first term 

\begin{align*}
   \left(\frac{ y_{0} + x_0L}{(L^2 - 1)x_0^2}  \right) \sum_j  \left(y_{j}x_0 -y_{0} x_j \right)^2 
\end{align*}

Close inspection of the second term 

\begin{align*}
 &  \sum_j [(y_{j} - \dfrac{y_{0}}{x_0} x_j)(\frac{x_j}{x_0})] \\
  & =\frac{1}{x^2_0} (x_0  \sum_j x_j y_j - y_0\sum_k x_j^2 ) \\  
   & = \frac{1}{x^2_0} (x_0 (L+x_0y_0) - y_0(x^2_0 -1) );  \text{Lorentz Dot product and Lorentz points} \\
   & = \frac{x_0 L + y_0}{x^2_0} 
\end{align*}

\begin{align*}
    & \sum_j  \left(y_{j}x_0 -y_{0} x_j \right)^2 - (L^2-1) \\
    & = \sum_j (y_j^2x_0^2 -2 x_0 y_0 x_j y_j + y_0^2 x_j^2) -((-x_0y_0 + \sum_k x_k y_k)^2-1) \\
    & = x_0^2 \sum_j y_j^2 - 2 x_0 y_0 \sum_j( x_j y_j) + y_0^2 \sum_j x_j^2 -(x_0^2 y_0^2 - 2 x_0 y_0 \sum_k( x_k y_k )+ (\sum_k x_k y_k)^2 - 1) \\
    & = x_0^2 \sum_j y_j^2 + y_0^2 \sum_j x_j^2 - x_0^2 y_0^2 - (\sum_k x_k y_k)^2 + 1 \\
    & = (1+ \sum_j x_j^2) \sum_j y_j^2 + (1+ \sum_j y_j^2) \sum_j x_j^2 - (1+ \sum_j x_j^2)(1+ \sum_j y_j^2) - (\sum_k x_k y_k)^2 + 1 ;\text{with} -x_0^2 + \sum x_j^2 = -1 \\
    & = \sum_j x_j^2 \sum_j y_j^2 - (\sum_j x_j y_j)^2; \textbf{remaining terms} \\
    & \geq 0 (Non-Negative values) ; \text{Cauchy Inequality}\\
    & > 0  (\text{non identical vectors})
\end{align*}

Finally, the sign of 
\begin{align*}
    sign(\nabla f^d \cdot (\nabla f^a)^T) & = sign(-\left(\frac{ y_{0} + x_0L}{(L^2 - 1)x_0^2}  \right) \left ( \sum_j  \left(y_{j}x_0 -y_{0} x_j \right)^2 - (L^2-1) \right ) ) \\
    & = sign((- x_0L) -y_0 ) ; x_0 >0, y_0 > 0, L \leq - 1 (\text{curvature of -1})
\end{align*}

For uncertain points, less gradient!! $x_0>y_0$ implies $ (- x_0L) > y_0$. When $ sign((- x_0L) -y_0)$ is $+ve$, it implies gradient synergy for the two loss components and faster convergence.  

\subsection*{Hyperbolic Coordinate Transformation}

Given:
\begin{itemize}
    \item $\mathbf{v} = [v_1, v_2, \ldots, v_k]$ is a vector
    \item $|\mathbf{v}| = \sqrt{\sum_{l=1}^k v_l^2}$ is the Euclidean norm
    \item $x_0 = \cosh(|\mathbf{v}|)$
    \item For $j = 1, \ldots, k$: $x_j = \sinh(|\mathbf{v}|) \cdot \frac{v_j}{|\mathbf{v}|}$
\end{itemize}

We want to compute $\frac{\partial x_j}{\partial v_l}$ for $j = 0, 1, \ldots, k$ and $l = 1, \ldots, k$.

\subsection*{Derivative of $x_0$ with respect to $v_l$}

\[
x_0 = \cosh(|\mathbf{v}|)
\]

Using the chain rule:
\[
\frac{\partial x_0}{\partial v_l} = \sinh(|\mathbf{v}|) \cdot \frac{\partial |\mathbf{v}|}{\partial v_l}
\]

We know:
\[
\frac{\partial |\mathbf{v}|}{\partial v_l} = \frac{v_l}{|\mathbf{v}|}
\]

Therefore:
\[
\frac{\partial x_0}{\partial v_l} = \sinh(|\mathbf{v}|) \cdot \frac{v_l}{|\mathbf{v}|}
\]

\subsection*{Derivative of $x_j$ with respect to $v_l$ for $j = 1, \ldots, k$}

\[
x_j = \sinh(|\mathbf{v}|) \cdot \frac{v_j}{|\mathbf{v}|}
\]

Let $u = \sinh(|\mathbf{v}|)$ and $w = \frac{v_j}{|\mathbf{v}|}$, so $x_j = u \cdot w$.

Using the product rule:
\[
\frac{\partial x_j}{\partial v_l} = \frac{\partial u}{\partial v_l} \cdot w + u \cdot \frac{\partial w}{\partial v_l}
\]

Compute $\frac{\partial u}{\partial v_l}$:
\[
\frac{\partial u}{\partial v_l} = \cosh(|\mathbf{v}|) \cdot \frac{\partial |\mathbf{v}|}{\partial v_l} = \cosh(|\mathbf{v}|) \cdot \frac{v_l}{|\mathbf{v}|}
\]

Compute $\frac{\partial w}{\partial v_l}$:
\[
w = \frac{v_j}{|\mathbf{v}|}
\]

Using the quotient rule:
\[
\frac{\partial w}{\partial v_l} = \frac{ \frac{\partial v_j}{\partial v_l} \cdot |\mathbf{v}| - v_j \cdot \frac{\partial |\mathbf{v}|}{\partial v_l} }{|\mathbf{v}|^2}
\]

We have:
\begin{itemize}
    \item $\frac{\partial v_j}{\partial v_l} = \delta_{jl}$ (Kronecker delta: 1 if $j = l$, 0 otherwise)
    \item $\frac{\partial |\mathbf{v}|}{\partial v_l} = \frac{v_l}{|\mathbf{v}|}$
\end{itemize}

So:
\[
\frac{\partial w}{\partial v_l} = \frac{ \delta_{jl} \cdot |\mathbf{v}| - v_j \cdot \frac{v_l}{|\mathbf{v}|} }{|\mathbf{v}|^2} = \frac{ \delta_{jl} }{|\mathbf{v}|} - \frac{v_j v_l}{|\mathbf{v}|^3}
\]

Combine all terms:
\begin{align*}
\frac{\partial x_j}{\partial v_l} &= \left( \cosh(|\mathbf{v}|) \cdot \frac{v_l}{|\mathbf{v}|} \right) \cdot \frac{v_j}{|\mathbf{v}|} + \sinh(|\mathbf{v}|) \cdot \left( \frac{ \delta_{jl} }{|\mathbf{v}|} - \frac{v_j v_l}{|\mathbf{v}|^3} \right) \\
&= \frac{v_j v_l}{|\mathbf{v}|^2} \cosh(|\mathbf{v}|) + \frac{\delta_{jl}}{|\mathbf{v}|} \sinh(|\mathbf{v}|) - \frac{v_j v_l}{|\mathbf{v}|^3} \sinh(|\mathbf{v}|)
\end{align*}

Factor terms:
\[
\frac{\partial x_j}{\partial v_l} = \frac{\delta_{jl}}{|\mathbf{v}|} \sinh(|\mathbf{v}|) + \frac{v_j v_l}{|\mathbf{v}|^2} \left( \cosh(|\mathbf{v}|) - \frac{\sinh(|\mathbf{v}|)}{|\mathbf{v}|} \right)
\]

\subsection*{Final Results}

\[
\boxed{
\begin{aligned}
\frac{\partial x_0}{\partial v_l} &= \frac{v_l}{|\mathbf{v}|} \sinh(|\mathbf{v}|) \\
\frac{\partial x_j}{\partial v_l} &= \frac{\delta_{jl}}{|\mathbf{v}|} \sinh(|\mathbf{v}|) + \frac{v_j v_l}{|\mathbf{v}|^2} \left( \cosh(|\mathbf{v}|) - \frac{\sinh(|\mathbf{v}|)}{|\mathbf{v}|} \right)
\end{aligned}
}
\]

for $j = 1, \ldots, k$ and $l = 1, \ldots, k$.

\textbf{Claim.} For $r>0$, 
\[
\cosh(r) - \frac{\sinh(r)}{r} > 0.
\]

\textbf{Proof.}
Define 
\[
g(r) = r\cosh(r) - \sinh(r).
\]
Then
\[
g'(r) = r\sinh(r) > 0 \quad \text{for } r>0,
\]
so $g$ is strictly increasing. Since $\lim_{r\to 0} g(r)=0$, we have $g(r)>0$ for all $r>0$. Dividing by $r>0$ gives the result.

\medskip
\textbf{Numerical check.}
Using Taylor expansions,
\[
\cosh(r)=1+\frac{r^2}{2}+O(r^4), \quad 
\sinh(r)=r+\frac{r^3}{6}+O(r^5),
\]
so
\[
\cosh(r) - \frac{\sinh(r)}{r}
= \frac{r^2}{3} + O(r^4) > 0 \quad \text{for small } r>0.
\]
\hfill $\square$

\end{document}